\documentclass[preprint,5p,compress]{elsarticle}

\usepackage{booktabs}
\usepackage{url}
\usepackage{amsmath}
\usepackage{stmaryrd}
\usepackage{xcolor}
\usepackage{multirow}

\journal{Neurocomputing}

\definecolor{mycolor}{rgb}{0,0,0} 
\newcommand{\highlight}[1]{{\color{mycolor}#1}}

\begin{document}

\begin{frontmatter}

\title{Tips, guidelines and tools for managing multi-label datasets: \\ the mldr.datasets R package and the Cometa data repository}

\author[UJA]{Francisco Charte\corref{cor1}}
\ead{fcharte@ujaen.es}

\author[UJA]{Antonio J. Rivera}
\ead{arivera@ujaen.es}

\author[UGR]{David Charte}
\ead{fdavidcl@ugr.es}

\author[UJA]{Mar\'ia J. del Jesus}
\ead{mjjesus@ujaen.es}

\author[UGR,IND]{Francisco Herrera}
\ead{herrera@ugr.es}

\cortext[cor1]{Corresponding author. \\ %Arxiv
  Manuscript accepted at Neurocomputing: \url{https://doi.org/10.1016/j.neucom.2018.02.011} \\
  \textcopyright~2018. This manuscript version is made available under CC  	BY-NC-ND 4.0 license \url{https://creativecommons.org/licenses/by-nc-nd/4.0/}
}
\address[UJA]{Department of Computer Science, University of Ja\'en, 23071 Ja\'en, Spain}
\address[UGR]{Department of Computer Science and A.I., University of Granada, 18071 Granada, Spain}
\address[IND]{Faculty of Computing and Information Technology, King Abdulaziz University, 21589, Jeddah, Saudi Arabia}

\begin{abstract}
New proposals in the field of multi-label learning algorithms have been growing in number steadily over the last few years. The experimentation associated with each of them always goes through the same phases: selection of datasets, partitioning, training, analysis of results and, finally, comparison with existing methods. This last step is often hampered since it involves using exactly the same datasets, partitioned in the same way and using the same validation strategy. In this paper we present a set of tools whose objective is to facilitate the management of multi-label datasets, aiming to standardize the experimentation procedure. The two main tools are an R package, mldr.datasets, and a web repository with datasets, Cometa. Together, these tools will simplify the collection of datasets, their partitioning, documentation and export to multiple formats, among other functions. Some tips, recommendations and guidelines for a good experimental analysis of multi-label methods are also presented. 
\end{abstract}


\end{frontmatter}

\section{Introduction}
The need to automatically label data has significantly increased in recent years, in line with the growth of multimedia content online, especially all types of social networks. \highlight{People and objects present in a photograph recently uploaded to Instagram or Facebook, subjects and areas related to an article published in a digital newspaper, or styles and emotions linked to a new melody} must be determined as quickly and accurately as possible. The large flow of new information published every minute on the Internet requires this functionality, essential to catalog each piece of data. This demand is satisfied by multi-label learning algorithms \cite{Charte:SB-MLC,TutorialVentura}, able to learn from prelabeled examples and then do this task automatically.

\begin{figure}
  \centering
  \includegraphics[width=\columnwidth]{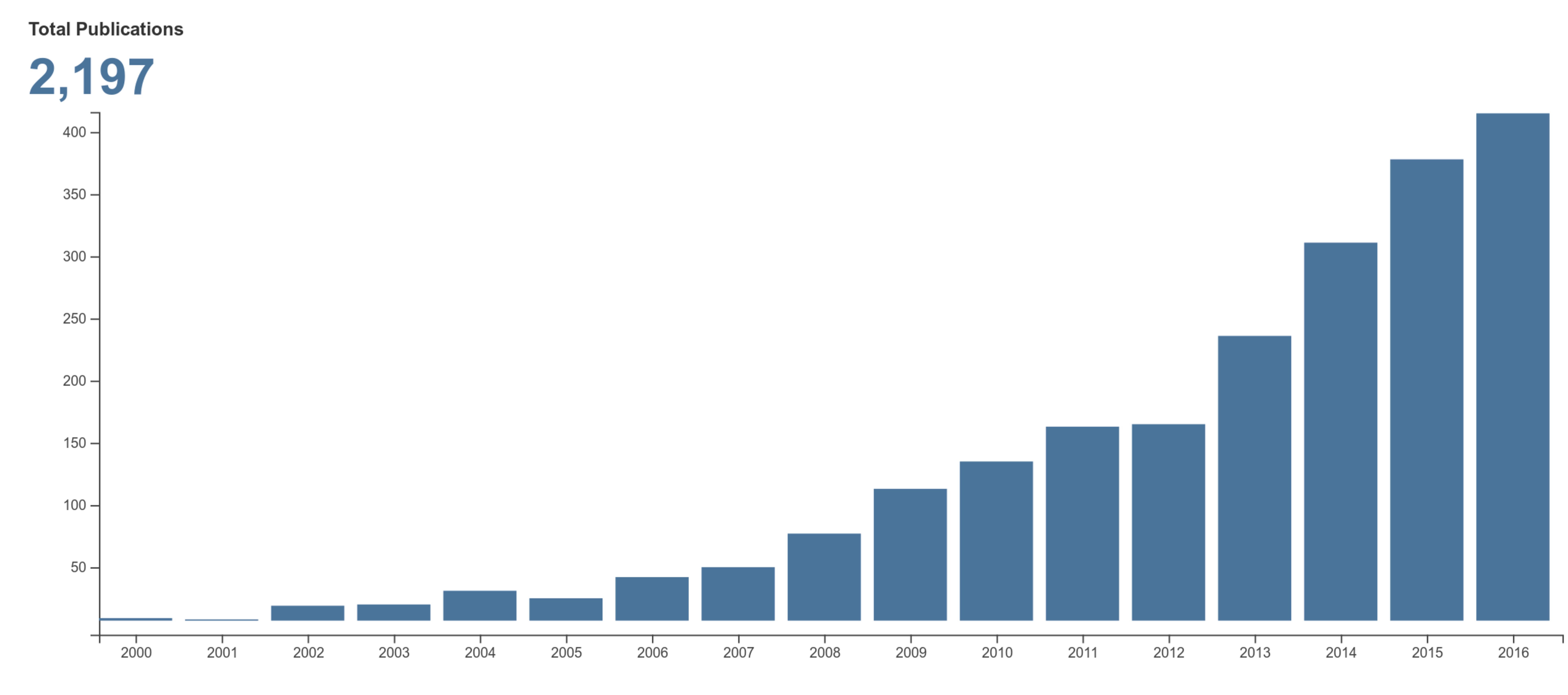}
  \caption{Number of published articles dealing with MLC 2000-2016 (source: Web of Science).}
  \label{Fig.MLCPapers}
\end{figure}

The knowledge obtained from these prelabeled data instances can be represented in disparate ways, i.e. decision trees \cite{Clare}, neural networks \cite{Zhang2}, support vector machines \cite{Elisseeff1}, etc. Since there are several labels associated to each data sample, the structure of these models tend\highlight{s} to be slightly more complex than is usual in traditional classification. Alternatively, there also exist certain label transformation techniques, such as binarization \cite{Godbole} and label powerset \cite{Boutell}, oriented to applying traditional classifiers to multi-label data. In addition, there are some very specific casuistries, such as imbalanced labels concurrence \cite{Charte:NeucomREMEDIAL} or high dimensionality both on the feature space \cite{spolaor2016systematic} and label space \cite{LI-MLC}. As a consequence\highlight{,} a plethora of multi-label classification (MLC) algorithms have been proposed lately, each of them claiming to perform better than the previous ones.

Proposing a new learning method implies comparing it with some existing algorithms. Doing so requires conducting an empirical experimentation. The experimental process customarily consists in the following steps:

\begin{enumerate}
    \item Collect some multilabel datasets (MLDs), analyze their traits to choose \highlight{those} most suitable to the task, and properly document them. Additionally, some data preparation steps may be performed, such as binarization.
    
    \item Run the proposed algorithm with the chosen data, and obtain a collection of performance indicators.
    
    \item Compare the indicators of the new method with those of some existing ones, so as to assess the proposal.
    
    \item Tune the algorithm as convenient and return to step 2 until it achieves a clear improvement.
\end{enumerate}

Although the process is apparently clear, accomplishing it in a proper way is not always straightforward. Practitioners frequently fail in some steps, drawing  conclusions of doubtful correctness. Sometimes the reason is in the scarcity of appropriate tools. Occasionally the obstacle is the lack of experience in such a specialized field.

As long-time MLC scholars we have developed several tools over the years to ease our work. In addition, we follow a standardized procedure for performing MLC experiments aiming to draw sound conclusions. Our goal in this paper is to present some of these tools, specifically the \texttt{mldr.datasets} R package and the comprehensive multi-label data source, Cometa. Furthermore, how to use these tools in order to avoid some of the pitfalls usually found during MLC experimentation is also explained.

\highlight{The main contributions in this paper can be summarized as follows:
\begin{itemize}
	\item We have identified the main pitfalls while performing multi-label experiments.
	
	\item A collection of good practices  aimed to overcome the previous traps is provided.
	
	\item We have developed a new tool, the \texttt{mldr.datasets} package introduced in Section \ref{Sec.mldr.datasets}, with the goal of easing multi-label data selection and preparation.
	
	\item By means of the previous tool, a comprehensive data repository has been generated. Cometa, presented in Section \ref{Sec.DataSource}, is a web application allowing to filter, search and select datasets, available in different file formats and partitioning schemes.
	
	\item A Docker image is provided to allow anyone to build their own multi-label data repository, automating the use of the previous tools.
\end{itemize}
}

The remainder of this paper is structured as follows. \highlight{The foundations about multi-label learning are provided in Section \ref{Sec.Background}.} Section \ref{Sec.MLCExperiments} describes how MLC experiments are usually conducted, while Section \ref{Sec.Tips} explains some of the frequent pitfalls and the way them can be surpassed. Section \ref{Sec.mldr.datasets} introduces the \texttt{mldr.datasets} R package. This tool has been used to build Cometa, the data repository presented in Section \ref{Sec.DataSource}. Lastly, Section \ref{Conclusions} provides the final conclusions.

\highlight{
\section{Multi-label learning background}\label{Sec.Background}
The main focus in this paper is put on the process to perform MLC and the tools needed to do it. However, MLC is part of broad field generically known as multi-label learning (MLL), where other kinds of tasks can be conducted as well. This section provides a brief introduction to MLL, a topic to which dozens of papers \cite{TutorialVentura} and even full books \cite{Charte:SB-MLC} have been devoted.

\subsection{MLL foundations}
Most common supervised machine learning tasks, such as binary and multiclass classification or regression, usually are guided only by an objective value. This would be the class assigned to a data pattern, in classification, or the target value to obtain as result of some kind of regression computation. Even methods which perform frequently non-supervised tasks, such as clustering, sometimes use this objective value to improve their results.

Multi-label learning \cite{Charte:SB-MLC} differs from the previous ones by the nature of the objective that guides the process. It is a set of binary values stating which labels are relevant to each data pattern, rather than a single class. Assuming $D$ is a dataset having $f$ input features, and being $L$ the full set of labels appearing in $D$, each data pattern would be constructed as shown in (\ref{Eq.Pattern}).

\begin{equation}
D_i = (X_i, Y_i) \mid X_i \in X^1\times X^2\times \dots\times X^f, Y_i \subseteq L
\label{Eq.Pattern}
\end{equation}

From this definition, mainly from the set of relevant labels (labelset) for each data sample ($Y_i$), it is easy to compute certain characterization metrics, as described in the following subsection.

\subsection{Characterization metrics}
Characterization measurements are useful to know traits of multi-label data, such as the multi-labeleness degree of a dataset, its imbalance level, the label sparseness, etc., thus being fundamental to choose the proper datasets for each case. The described below are among the most used ones.

\paragraph{Label cardinality}
    It is defined in \cite{Charte:SB-MLC} as shown in (\ref{Eq.Card}), where \textit{D} is any MLD, \textit{n} the number of \highlight{instances}, \textit{k} the number of labels, and $Y_i$ the labelset corresponding to the i-th data sample. \textit{Card} is the average number of relevant labels (number of labels active per instance) for the MLD \textit{D}.

\begin{equation} 
\mathit{Card}\left(D\right) = \frac{1}{n} \displaystyle\sum\limits_{i=1}^{n} \lvert Y_i\rvert 
\label{Eq.Card}
\end{equation}

\paragraph{Label density}
	It is defined as (\ref{Eq.Dens}). Usually, \textit{Card} changes along with the total number of distinct labels. So a normalized version, named \textit{Dens}, is defined as \textit{Card} divided by the total number of labels.

\begin{equation}
\mathit{Dens}\left(D\right) = \frac{1}{k} \frac{1}{n} \displaystyle\sum\limits_{i=1}^{n} \lvert Y_i\rvert \label{Eq.Dens}
\end{equation}

\paragraph{meanIR}
	This measure is computed as (\ref{Eq.MeanIR}) the average imbalance ratio of each label, the \textit{IRLbl} (\ref{Eq.IRLbl}). In these equations \textit{L} stands for the full set of labels appearing in the MLD. Both measures were introduced in \cite{Charte:Neucom13} to assess the imbalance level in an MLD.

\begin{equation}
\textit{MeanIR} = \frac{1}{k} \displaystyle\sum\limits_{l \in L}^{}\textit{IRLbl}(l) .
\label{Eq.MeanIR}\end{equation}

\begin{equation} \textit{IRLbl}(l) = 
\frac{
\displaystyle\max\limits_{l' \in L}^{}
\left(\displaystyle\sum\limits_{i=1}^{n}{\llbracket l' \in Y_i \rrbracket}\right)
}
{
\displaystyle\sum\limits_{i=1}^{n}{\llbracket l \in Y_i \rrbracket}} . 
\label{Eq.IRLbl}
\end{equation}

\paragraph{SCUMBLE and SCUMBLE.CV} 
	These two metrics are aimed to evaluate the level of concurrence among minority and majority labels. Introduced in \cite{Charte:NeucomSCUMBLE}, the former is defined (\ref{Eq.SCUMBLE}) as the average \textit{SCUMBLE} of each instance (\ref{Eq.SCUMBLEins}) in the dataset. The latter is simply the coefficient of variation associated to this average.

\begin{equation}
\textit{SCUMBLE}\left(D\right) = \frac{1}{n}
\displaystyle\sum\limits_{i=1}^{n} \textit{SCUMBLE}_{i}
\label{Eq.SCUMBLE}
\end{equation}

\begin{equation}
\textit{SCUMBLE}_{i} =
1 - \frac{1}{\overline{\textit{IRLbl}_i}}\left(\prod\limits_{l \in L}^{} \textit{IRLbl}_{il}\right)^{\left(1/k\right)}
\label{Eq.SCUMBLEins}
\end{equation}

\paragraph{TCS}
	This metric (\ref{TCS}) was presented in \cite{Charte:HAIS16} as a straightforward way to assess the theoretical complexity of an MLD. It is based on just three traits of the dataset: $f$ stands for the amount of input features, $k$ for the number of labels, and $ls$ is the total number of label combinations in $D$. The larger is the value returned by this measurement, the harder would be to learn a predictive model from the dataset.

\begin{equation}
\textit{TCS}(D) = \log(f \times k \times ls ) \label{TCS}
\end{equation}

All these metrics can be easily obtained through the \texttt{mldr.datasets}, as described in Section \ref{Sec.mldr.datasets}.

\subsection{Main MLL tasks}
Aside from performing exploratory data analysis, different machine learning tasks can be faced while working with multi-label data. The following are among the most usual ones:

\paragraph{Classification} 
It is arguably the most studied problem in multi-label learning. The goal is to find a model able to predict the labelset for new data patterns. As described in \cite{Charte:SB-MLC}, two main approaches to model a new classifier exist. The first one aims to transform the original data so that traditional classifiers can be used. The two main transformation techniques are known as BR (\textit{Binary Relevance}) \cite{Godbole} and LP (\textit{Label Powerset}) \cite{Boutell}. The former deals with each label separately, using a set of binary classifiers to make the prediction. while the latter joins the labels to create a class identifier, relying in a multiclass classifier as predictive model. The second approach consists in adapting existing classification methods to handle multi-label patterns, instead of transforming them. The use of ensembles is also very popular in the field, with proposals as ECC (\textit{Ensemble of Classifier Chains}) \cite{Read}, EPS (\textit{Ensemble of Pruned Sets}) \cite{Read:2008:2}, etc.  
	
\paragraph{Ranking} 
As the name denotes, label ranking methods are used to elaborate a ranking of labels according to their relevance for a data pattern. Therefore, instead of producing a labelset, a string of 0s and 1s stating which labels are predicted  as classifiers do, these methods assign a weight to each label. This label ranking can be used directly, as well as transformed into a predicted labelset by applying a specific threshold. There are many proposed label ranking methods, RPC (\textit{Ranking by Pairwise Comparison}) \cite{RPC} and CLR (\textit{Calibrated Label Ranking}) \cite{CLR} are two of the best known.

\paragraph{Clustering} 
Usually, clustering methods \cite{ClusteringSurvey2015} work in an unsupervised manner. Therefore, there would be no difference between clustering binary, multiclass or multi-label data. However, sometimes class information is taken into account to improve clustering results. Both basic and hierarchical clustering have been used as tools to create predictive multi-label methods. For instance, the ML-RBF algorithm \cite{Zhang4} relies on the classic k-means algorithm to cluster the data points and use the clusters as centers of the radial basis functions in the hidden layer. The HOMER algorithm \cite{HOMER} produces an hierarchical model by clustering the patterns in each node, introducing the concept of \textit{meta-label} to represent similar labels. Only a few multi-label specific clustering methods have been proposed until now. One of them \cite{MLC-Clustering} is an evolutionary algorithm able to perform distance metric learning. The method considers multiple labels per cluster, computing a cluster validity measure from the relationships among neighbors. In \cite{DensityClustering} a density-based algorithm, similar to DBSCAN \cite{DBSCAN}, is proposed as potential solution to perform multi-label clustering.

\subsection{MLL evaluation metrics}
The metrics used to evaluate a result depend on the performed task, but also on the own nature of the analyzed data. While in binary classification \textit{Accuracy} or \textit{Precision} are the most usual performance measures, and there are only a handful more to choose from, when the class of patterns is not binary but multi-label the group of available metrics is considerably larger. The difference is that the output of a binary classifier is either correct or incorrect, while that of a multi-label can be totally or partially correct. This justifies the existence of around twenty metrics for multi-label classification only, as explained in \cite[Chapter~3]{Charte:SB-MLC}.

A multi-label classifier outputs a bipartition as result. That is a sequence or array of 0s and 1s, stating which labels are predicted as relevant for a data sample and which not (the predicted labelset). These predictions are aggregated to produce several confusion matrices, from which the usual classification performance metrics can be computed. Depending on how the aggregation is conducted, the metrics are grouped into two large categories: example-based and label-based metrics.

Example-based measurements are computed individually from each data pattern in the evaluated set. These values are then averaged, simply dividing by the number of evaluated samples. Some of the  most usual metrics in this group are \textit{Hamming loss}, \textit{Accuracy}, \textit{Precision}, \textit{Recall} and \textit{F-measure}. Unlike the other ones, \textit{Hamming loss} (\ref{Eq.HL}) is not common in traditional classification. Being $n$ the number of data points and $k$ the number of considered labels, it calculates the symmetric difference ($\Delta$ operator) between the predicted labelset $Y_i$ and the ground truth $Z_i$, thus counting the number of mismatches. Therefore, it is a performance metric to be minimized instead of maximized.

\begin{equation}
\mathit{Hamming~loss} = \frac{1}{n} \frac{1}{k} \displaystyle\sum\limits_{i=1}^{n} \left\lvert Y_i \Delta Z_i\right\rvert  \label{Eq.HL}
\end{equation}

Instead of mixing the results of all labels in each instance, those can be separately aggregated and the evaluation metrics computed for each label. This is the way label-based performance metrics work. In fact, there are two ways to perform the averaging, as shown in equations (\ref{Eq.MacroB}) and (\ref{Eq.MicroB}). The macro-averaging approach sums the number of true positives, false positives, true negatives and false negatives for each label, and independently computes the measurement for each label. Thus the metric, such as \textit{Precision}, \textit{Recall}, \textit{F-measure}, etc., is calculated several times, as many as labels there are. Lastly, these measurements are added and divided by the number of labels ($k$). By contrast, in the micro-averaging approach the metric is computed only once, after the counters for all label have been aggregated.

\begin{equation}\small
  \mathit{Macro~metric}= \frac{1}{k} \sum\limits_{l \in \mathcal{L}}\mathit{EvalMet}\left(\textit{TP}_l,\textit{FP}_l,\textit{TN}_l,\textit{FN}_l\right) \label{Eq.MacroB}
\end{equation}
\begin{equation}\small
\mathit{Micro~metric} = \mathit{EvalMet}\left(\sum\limits_{l \in \mathcal{L}}^{}\textit{TP}_l,\sum\limits_{l \in \mathcal{L}}^{}\textit{FP}_l,\sum\limits_{l \in \mathcal{L}}^{}\textit{TN}_l,\sum\limits_{l \in \mathcal{L}}^{}\textit{FN}_l\right) \label{Eq.MicroB}
\end{equation}

The third main group of multi-label evaluation metrics is aimed to work over label rankings, instead of bipartitions. \textit{One error}, \textit{Ranking loss}, \textit{Coverage} and \textit{Average precision} are among the best known metrics in this category. These metrics usually check if a true relevant label is in the ranking produced by the algorithm, the number of steps to walk until a relevant label is found in the ranking, or whether a non-relevant label has been ranked above a relevant one. They can be computed even when the used algorithm produces a bipartition instead of a ranking, by relying in some other kind of real value as can be the confidence or a set of weights in a neural network.

Aside from these three main groups, some more specific multi-label metrics  have been defined to evaluate hierarchical \cite{MLC-Hierarchical} multi-label classification or the quality of multi-label clustering \cite{MLC-Clustering}.
}

\begin{figure*}
  \centering
  \includegraphics[width=\textwidth]{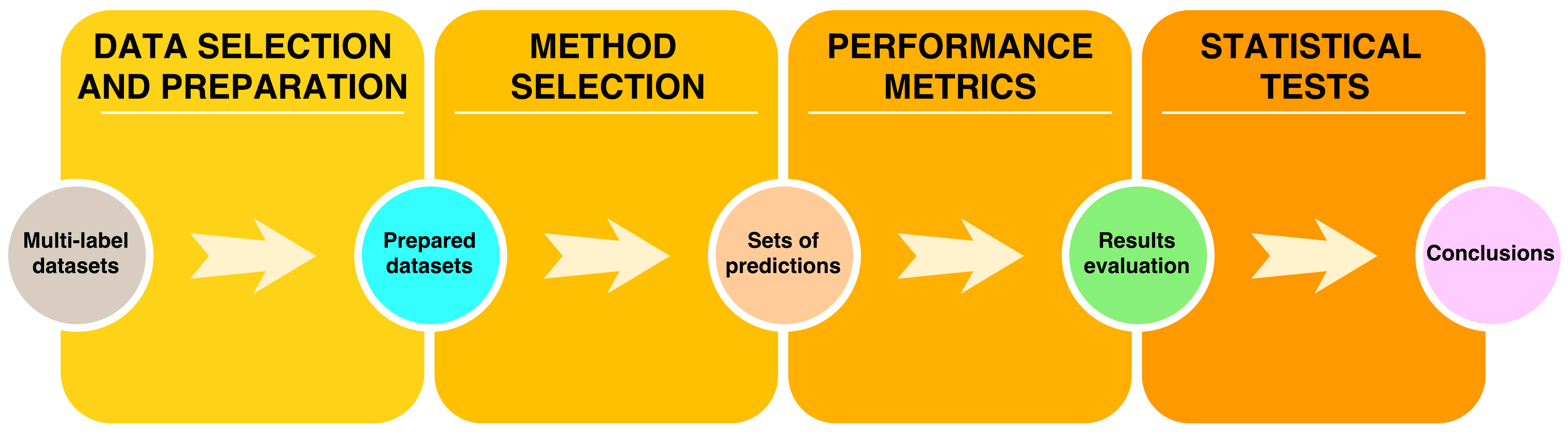}
  \caption{Steps in a typical MLC experiment.  \highlight{1) Datasets have to be selected from the available repositories, taking into account their traits in order to choose the more appropriate ones for the task at hand. Some data preparation could be needed, depending on possible data peculiarities (i.e. an imbalanced dataset could be balanced through a resampling algorithm). 2) Methods related to the proposed one have to be selected and run, obtaining a set of predictions from each one of them. 3) These predictions have to be evaluated, picking from the large set of available performance metrics those more adequate to the analyzed problem. 4) It is advisable to draw conclusions including statistical tests over the previous results.}}
  \label{Fig.Process}
\end{figure*}

\section{Conducting multi-label learning experiments}\label{Sec.MLCExperiments}
Let us assume we are designing a new MLC algorithm aimed to improve the results produced by existing ones. Aside from explaining the theoretical \highlight{hypotheses} underlying the new proposal, stating why it should perform better, \highlight{we also commit to providing} real evidence of this enhancement. Therefore, an empirical study has to be conducted.

The main steps usually followed to carry out a data mining experiment are the four ones previously enumerated, also depicted in Fig. \ref{Fig.Process}. Here we are delving into some specific aspects. Later, we will put the focus on the usual traps and on how to surpass them.

\subsection{Data selection and preparation}
The first step in this process is usually the selection of datasets to be used in experiments. The criteria employed to select the data that will be involved in the experimentation may vary. In the MLC context aspects such as the number of distinct labels, label cardinality and density, imbalance levels and label concurrence, among others, are usually taken into consideration.

Depending on the case, these data can be either real or synthetic. The use of data collected from actual sources has multiple advantages. In this way, the tested algorithm will be exposed to the characteristics and complexities of real-world information, in the same context in which it is presumably intended to be used. However, sometimes data of this kind do not meet the needs of ongoing experimentation or do not fit the specific traits required. In these cases specialized tools, capable of generating data that match the required characteristics, tend to be the solution.

Once the datasets to be included in the experimental study have been collected, it is necessary to apply the preprocessing steps that are considered appropriate. For instance, a data transformation technique such as label powerset or binary relevance, would allow to apply a multiclass or binary classifier. Moreover, training and test partitions have to be extracted from the original MLD. Sometimes, depending on the algorithm, the training set can be also divided into two subsets, using one of them to train the model and the other to validate its parameters.

\subsection{Competing methods}
One of the objectives of proposing a new learning me\-thod is to improve the results of others. Although sometimes the algorithm presented may be completely new, in most cases it will be an improvement over a method already in use. In any case, it is to be assumed that the authors have certain knowledge of the field under study and, in particular, of methods similar to their own.

A large portion of MLC algorithms are based on transformation techniques, such as binarization \cite{Godbole} and label powerset \cite{Boutell}. Those allow to transform a complex problem (MLC) into several easier ones, but at the expense of increasing the computational costs consumed to process them. Therefore, running a large set of MLC methods over several datasets can be very time consuming, thus the importance of choosing the proper ones.

\highlight{
The alternative to the previous transformation techniques are classification methods adapted to deal with raw multi-label data. A plethora of proposals have been presented in this field, including multi-label decision trees \cite{Clare}, instance-based classifiers \cite{Zhang1}, neural networks \cite{Zhang2} and support vector machines \cite{Elisseeff1}, among other adaptations of traditional classification methods. 

Deep learning techniques have also found their niche in the classification field. Different deep neural networks architectures \cite{DeepNNsArchitectures} have been proposed in late years, and several of them, including DBNs (\textit{Deep-Belief Networks}) and CNNs (\textit{Convolutional Neural Networks}) have been proposed to classify multi-label data \cite{ReadDBN,DL-MLC-Land,CNN-MLC}. The superior performance of most deep learning methods is due to the integrated feature learning phase, able to extract a reduced set of new, more informative features. This task can be conducted as a preprocessing phase, for instance by relying on autoencoders \cite{Charte:AEReview}, then applying any multi-label classifier over the reduced feature set.
}

Once the methods that are going to compete with the new one are selected, attention must be paid to the configuration parameters \highlight{present in} each one of them. These should adjust to the recommendations given by their authors, otherwise their behavior could be \highlight{unexpected}. This is important only if we intend to run such algorithms, rather than take the published results. However, in the latter case, other aspects need to be taken into account as indicated below.

\subsection{Performance metrics}\label{Sec.PerformanceMetrics}
The selection of performance metrics is another key factor in the experimental process, specially in the MLC field. Unlike traditional classification, where usually a pair of performance indicators such as precision or accuracy tend to be enough, more than twenty evaluation measures are of common use in MLC. That is why picking the right ones for each case can usually have a large impact in the final conclusions.

As the compilation of performance metrics provided in \cite{Charte:SB-MLC} shows, the assessment of successes and failures can originate in a bipartition matrix or in a ranking. Moreover, the number of hits and misses can be aggregated following different strategies, by label or by sample.

It is obvious that the same performance metrics must be obtained for all the methods involved in the experimentation, otherwise it would not be possible to make a correct evaluation of the results. In addition to performing a detailed comparison, method against method and measure by measure, it is also necessary to obtain an overall appraisal for these comparison results. To accomplish this task it is usual to rely on the proper statistical tests.

\section{Tips and pitfalls while performing multi-label experiments}\label{Sec.Tips}
Any new algorithm proposal should begin with a review of existing literature on related methods, detailing their similarities, strengths and weaknesses, etc. Authors should establish the niche they intend to occupy with their proposal or the direction in which they want to improve the methods already published. Hereafter, the experimentation is usually carried out.

Most of the mistakes made during experimentation involve the data used and how \highlight{they are treated}. You must select the appropriate datasets for the task \highlight{at} hand and the proposed algorithm type. In addition, such data should be prepared in line with the experiments already published on related methods. Occasionally there are obstacles that hinder the correct use and preparation of the data. This is why in the following sections we focus mainly on this aspect, and why in sections 4 and 5 we present tools to overcome these obstacles.

Logically, the rest of aspects mentioned above, such as the selection of methods to be compared against the proposal, the set of metrics used to evaluate their performance and the usage of statistical tests, are also important. They will therefore be addressed later, albeit more briefly.

\subsection{Selecting the proper datasets}
While designing a new algorithm, sometimes the aim is to make it a general purpose method, with the goal of improving prediction in a broad way. However, in other cases the proposal is more specific. In the multi-label field, it is common to present methods tailored for dealing with large quantities of labels, with data showing imbalance among labels, with missing labels, etc. Each scenario requires a specific type of dataset to perform the experiments.

The basic principle must be that the data used in experimentation should present the problem for which the algorithm is proposed. This, however, is not always the case. There are authors who propose a new method to deal with large sets of labels, but in their experimentation they use MLDs that only have a few dozen of them. Logically, to support the behavior of the proposed method, it must be demonstrated that it works correctly with the right configuration: against datasets having hundreds or thousands of labels. This scenario can be extrapolated to similar ones: if a method attempts to address multi-label imbalance, the selected datasets should present this problem (very common in this field, on the other hand)\highlight{; there would be} no point in experimenting using only balanced datasets.

When choosing datasets, it would be advisable to observe the following recommendations:
\begin{itemize}
	\item Begin by performing an exploration of the characteristics of the available datasets, obtaining metrics that allow you to know the number of labels, cardinality, degree of imbalance, level of concurrence between labels, etc. All this information is vital to be able to select those sets that best suit your needs.
	
	\item Which datasets have other authors used to address the same task? \highlight{By answering} this question, a list of datasets commonly included in similar experiments can be obtained. It is a work that can be done at the same time as reviewing related works. It will also make it easier to compare results with previously published methods.
\end{itemize}

If the new method is presented as a generic multi-label classifier, datasets with \highlight{characteristics as diverse} as possible should be included in the experimentation. This approach will make it possible to identify the strengths and weaknesses of the proposal, key aspects when comparing it with existing methods. Again, the selection of such datasets should be based on an exploration of the characteristics of all publicly available ones.

\subsection{Preparation of data and methods}
The main objective of a multi-label experimentation is to compare a new algorithm with existing ones. For this, it is essential that the same training data are used in all cases, since the model obtained depends fundamentally on this condition. Aspects such as the rate of committed errors, the classifier's bias towards certain classes, its ability to generalize, etc., are highly influenced by the samples used during training. 

During our years in the MLC research field we have had the opportunity to find, in relation to data preparation and processing, the following mistakes:

\begin{itemize}
    \item Comparing the results obtained by running the proposed algorithm with those previously published for other methods. Unless exactly the same datasets have been used in the experimentation, with exactly the same pre-processing and partitioning strategy, such comparison will not be valid. Taking the performance measures published in an article for a certain method is highly convenient, but if you train that same method using unidentical partitions for training and testing the results will differ.
    
    \item Delivering complete datasets to each method and allowing them to be partitioned internally. Many tools automate random partitioning when evaluating an algorithm. However, in this situation, the exact same training partitions would not be used to generate the models, which would induce dissimilarities that benefit some and harm others.
    
    \item Running the algorithms to be compared but using different datasets in some cases. For example, using hold-out with certain methods because they are slower and cross validation with others, or reducing the set of input attributes for some methods and not for others. Obviously in these cases the authors are artificially favoring some algorithms over others.
\end{itemize}

For a fair comparison of several classifiers, it is therefore essential to train the models with exactly the same data samples. Starting from this premise, from our point of view there are the following alternatives:

\begin{itemize}
    \item If you want to compare one algorithm with the published results of another, without the need to run the latter, it is essential to have the data partitions used by its authors. This is only possible if, together with the results of such method, these partitions are also made publicly available. This approach is complicated as soon as it becomes necessary to compare with two or more already published methods, since different authors will have used disparate data partitions. In this situation the new algorithm should be run with the appropriate data for each case and pairwise comparisons should be made. In our opinion it is the least advisable alternative. It should only be used when it is impossible to run an existing method, but the data used to evaluate its performance is available.
    
    \item In any other case, without having the data originally used by other authors, the procedure to follow is always the same. The data must first be prepared, including partitioning, so that all algorithms to be tested  receive the same set of training and testing samples. Then, the code of all methods compared should be obtained, preferably from their authors. If this is not possible, they should be implemented as accurately as possible from the article in which they are described. Finally, all the algorithms must be executed using the same data and obtaining the corresponding predictions.
\end{itemize}

Trying to follow these instructions can easily lead to a number of obstacles. If you have access to the original data, its format may not be appropriate for the tools used to implement the new algorithm. In fact, almost every time several methods are going to be run, coming from different authors, there is diversity in file formats. These are the kind of issues the tools described in Sections \ref{Sec.mldr.datasets} and \ref{Sec.DataSource} strive to overcome.

\subsection{Assessing algorithm performance}
As stated in Section \ref{Sec.PerformanceMetrics}, a multi-label classifier can be evaluated using a wide range of performance metrics. Each of them offers a different quality indicator, so it is essential to use a good set of them to obtain the most balanced possible assessment.

If the experimentation includes results taken directly from previous publications, and bearing in mind the assumptions indicated in the previous sections, then the selection of measures will be given to us. It will be necessary to use the metrics already used in these publications, not others. The results obtained by the proposed method should therefore be used to calculate these measurements, thus allowing direct comparison with the algorithms already published.

Even if we have exactly the same data partitions to train our model, as explained above, and calculate the same set of measures, the comparison with other algorithms may not be completely fair. The calculation of some of the multi-label performance metrics is relatively complex, and there may be differences in the way they are computed between different multi-label software packages.  Consequently, the only way to be absolutely certain that the results are comparable would be to ensure that the computation of the measures is also carried out in the same way. 

If we are going to run all the algorithms ourselves, we will have total freedom in choosing the performance metrics. In this case\highlight{, as recommended in \cite{Madjarov},} we should select a broad set of evaluation metrics, including both sample-based and label-based measures, as well as measures calculated on bipartite and label rankings. \highlight{In particular, it is advisable to include at least one or two metrics from each of the following groups:
	
	\paragraph{Sample-based metrics based on bipartition results} As explained in Section \ref{Sec.Background}, these metrics are computed sample by sample and then an average evaluation is obtained. \textit{Hamming loss} is maybe the most common metric in this group; \textit{Accuracy}, \textit{Precision}, \textit{Recall} and \textit{F-measure} being popular as well. Most of them are obtained from the confusion matrix for each label. We recommend including always \textit{Hamming loss} and \textit{F-measure}, since the former one is the complement of \textit{Accuracy} and the latter is the harmonic mean of \textit{Precision} and \textit{Recall}. 
	
	\paragraph{Metrics based on ranking results} Although many algorithms produce a bipartition as result, stating which labels are predicted for the data patterns, many others provide a label ranking. In this case, the set of predicted labels is obtained after applying a certain threshold over this ranking. Ranking-based metrics evaluate the performance operating with a label ranking, being \textit{One error}, \textit{Ranking loss} and \textit{Average precision} among the most frequently used.
	
	\paragraph{Label-based metrics} Most metrics obtained from a confusion matrix, such as \textit{Precision} and \textit{Recall}, can be computed by label instead of by sample, as described in Section \ref{Sec.Background}. Since two averaging strategies exist, named \textit{macro-averaging} and \textit{micro-averaging}, two measurements can be retrieved for most of these metrics. In our opinion,  \textit{Macro F-measure} and \textit{Micro F-measure} should be included in most evaluations as they provide two additional views on method performance.

The selection of evaluation metrics } may be influenced if the proposed algorithm addresses a specific problem. For example, if the analysis corresponds to methods for working with imbalanced data, metrics such as \textit{AUC} or \textit{Macro F-measure} would be more appropriate than \textit{Hamming loss} or \textit{Accuracy}, as they are less biased towards majority labels. 

\begin{table*}[htp!]
	\setlength{\tabcolsep}{0.5em}
	\def\arraystretch{0.9}
	\centering
	\highlight{
		\footnotesize
		\begin{tabular}{p{.17\textwidth}p{0.39\textwidth}p{0.38\textwidth}}
			
			\toprule
			\textbf{Stage} & \textbf{Avoid...} & \textbf{Instead, ...}\\
			\midrule
			Data selection &
			using datasets which not present the tackled problem. &
			explore the characteristics of available datasets for proper selection. \vspace{12pt} \\  &
			skipping popular datasets for the task. &
			review related works and look for recurring datasets. \\
			\midrule
			Method preparation &
			comparing your results to previously published ones. &
			perform new runs of each competing method under same conditions. \vspace{12pt}\\ &
			partitioning datasets differently for each method. &
			build partitions prior to running methods. \vspace{12pt}\\ &
			performing different validations for each method. &
			determine the validation strategy taking into account possible slow methods. \\
			\midrule
			Algorithm assessment &
			choosing metrics that dismiss the tackled problem. &
			select those which are affected by it. \\
			\midrule
			Conclusions &
			hiding the behavior of a method behind average values. &
			rank methods according to their performance. \vspace{12pt}\\ &
			performing statistical tests without guaranteeing their assumptions. &
			possibly compare methods with non-parametric statistical tests. \\
			\bottomrule
		\end{tabular}
	}
	
	\caption{\highlight{General tips on how to avoid common mistakes during an experimentation in the multi-label field.}}
	\label{Table.Tips}
\end{table*}

\subsection{Reaching global conclusions}
When an experimental study is conducted involving multiple datasets, several algorithms and various performance metrics, the number of indicators to be evaluated is so large that it may be difficult to reach an overall conclusion. Usually, a classifier will be better than others while working with certain MLDs, but its behavior will worsen with other datasets. The same is applicable to the use of several evaluation metrics.

A first approach to a global assessment could be to count the number of times each algorithm wins or loses against the rest. From here, a ranking to determine the position of each method, usually grouped by performance metric, is almost immediate. In our opinion this approach is better than others we have found sometimes in the literature, such as averaging all the results from each algorithm. One of them can be the best one when evaluated with a certain dataset or metric, but perform horribly in all other cases. The authors should strive to show a realistic perspective of its behavior, for instance by means of a ranking instead of an average, which would hide it.

The conclusions of an experimental analysis can be reinforced by using more formal procedures, carrying out the relevant statistical tests. In most cases it is not possible to guarantee conditions of normality (the results to be evaluated following a normal distribution) and homoscedasticity (the variance being homogeneous). For this reason, non-parametric statistical tests are commonly used. Depending on the number of methods to be compared, and if such comparison is pairwise or multiple, the proper tests have to be chosen as explained in \cite{Garcia:2008,Salva:INS2010}.

\highlight{
\subsection{Summary, advantages and disadvantages}
To make it easier to follow up on previous advice, it has been summarized in Table \ref{Table.Tips}. The first column indicates the stage of the process, second the trap to avoid and third one what to do instead. In our opinion, adhering to these recommendations would bring several advantages to any MLL study:

\begin{itemize}
    \item The selection of the proper datasets, those that presumably present the  problem that the proposed me\-thod aims to solve, for instance a high imbalance level, will back the behavior of the method.
    
    \item Correctly comparing our results with published ones, either by using the same data partitions or by running the other methods with our data, will make the study more solid.
    
    \item Choosing a good set of evaluation metrics, specially those designed to expose strengths and weaknesses of a method while facing specific problems, will be more convincing to other researchers. This confidence in the results can be increased if the appropriate statistical tests are conducted.
\end{itemize}

Faced with these benefits, the main disadvantage in following these tips is the increased amount of work to be done. Time has to be devoted to explore  dataset traits, in order to choose the best ones, as well as to obtain data partitions and other methods implementations. Moreover, finding the tools to tackle all this work is not always easy. Aiming to mitigate this last obstacle,  the tools we have developed to overcome these tasks are introduced in the following sections.

}

\section{mldr.datasets: The tool for managing multi-label datasets}\label{Sec.mldr.datasets}
As the previous section has shown, selecting and preparing datasets to be used for experimentation are essential steps. However, when it comes to obtaining the appropriate datasets, partitioning them and obtaining them in the right format for each learning algorithm, multiple obstacles arise. It would be desirable to have a tool that facilitates such operations, as well as easing the exploration of its characteristics, being able to provide the reference for each dataset, etc. That is the motivation behind the development of the \texttt{mldr.datasets} package.

The first version of this software package for R users was introduced in \cite{Charte:RUMDR}. Since then, its functionality has been extended by including new multi-label partitioning algorithms, new functions to export the data to disparate formats, automatic checking for data sparsity, etc. The main goal has been to facilitate all the tasks needed to select and prepare MLDs for conducting a experimentation.

This section provides a didactic description of the above mentioned package, explaining how to complete each of the tasks from obtaining a set of data to its exploration, documentation, partitioning and export.

\subsection{Installing mldr.datasets in our computer}
The \texttt{mldr.datasets} software is an R package. As a consequence, anyone interested in using it needs to have the R interpreter \cite{RProject} installed in their computer. Assuming that this is the case, the installation procedure is the same followed for any package available in CRAN (\textit{Comprehensive R Archive Network}), to issue the following command at the R console:

\begin{quote}
\texttt{> install.packages("mldr.datasets")}
\end{quote}

This will install the last stable version of the package, which is 0.4 at this time. Development versions, with added functionality, are available in GitHub\footnote{The source code of the package is publicly available at \url{github.com/fcharte/mldr.datasets}.}. Assuming that the \texttt{devtools} package \cite{devtools} is installed and loaded, the most recent version of \texttt{mldr.datasets} can be always installed from GitHub as shown below:

\begin{quote}
\texttt{> install\_github("fcharte/mldr.datasets")}
\end{quote}

\begin{figure}[h!]
  \centering
  \fbox{\includegraphics[width=.9\columnwidth]{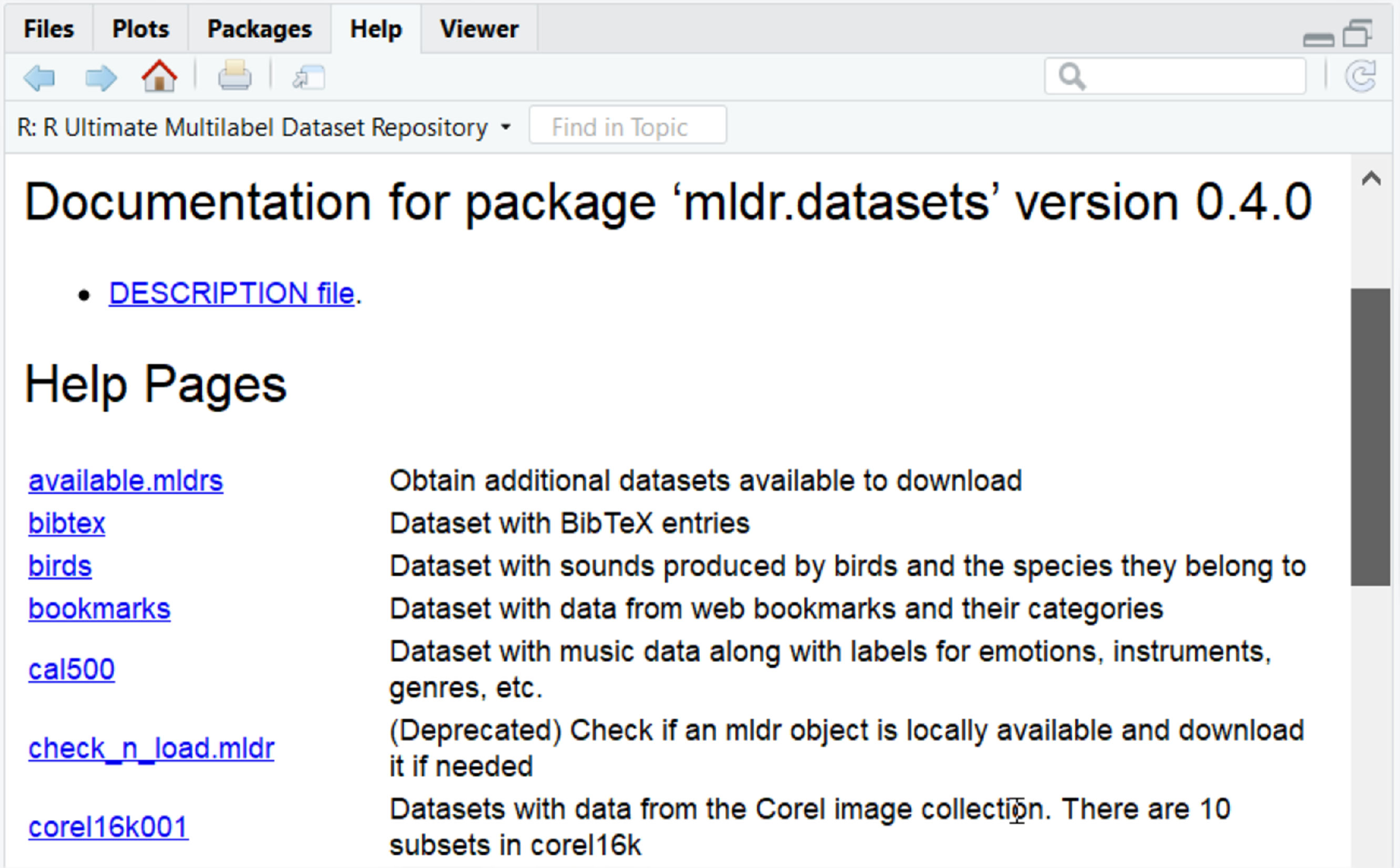}}
  \caption{Help index of the mldr.datasets package.}
  \label{Fig.Help}
\end{figure}

Once installed, the package has to be loaded into memory each time a new R session is started. This can be done with the usual \texttt{library()} or \texttt{require()} R commands. Since this moment the user can access the functions provided by the package. The index of all available functions (partially shown in Fig. \ref{Fig.Help}) can be retrieved with the following command:

\begin{quote}
\texttt{> help(package="mldr.datasets")}
\end{quote}

\begin{figure}[h!]
  \centering
  \fbox{\includegraphics[width=\columnwidth]{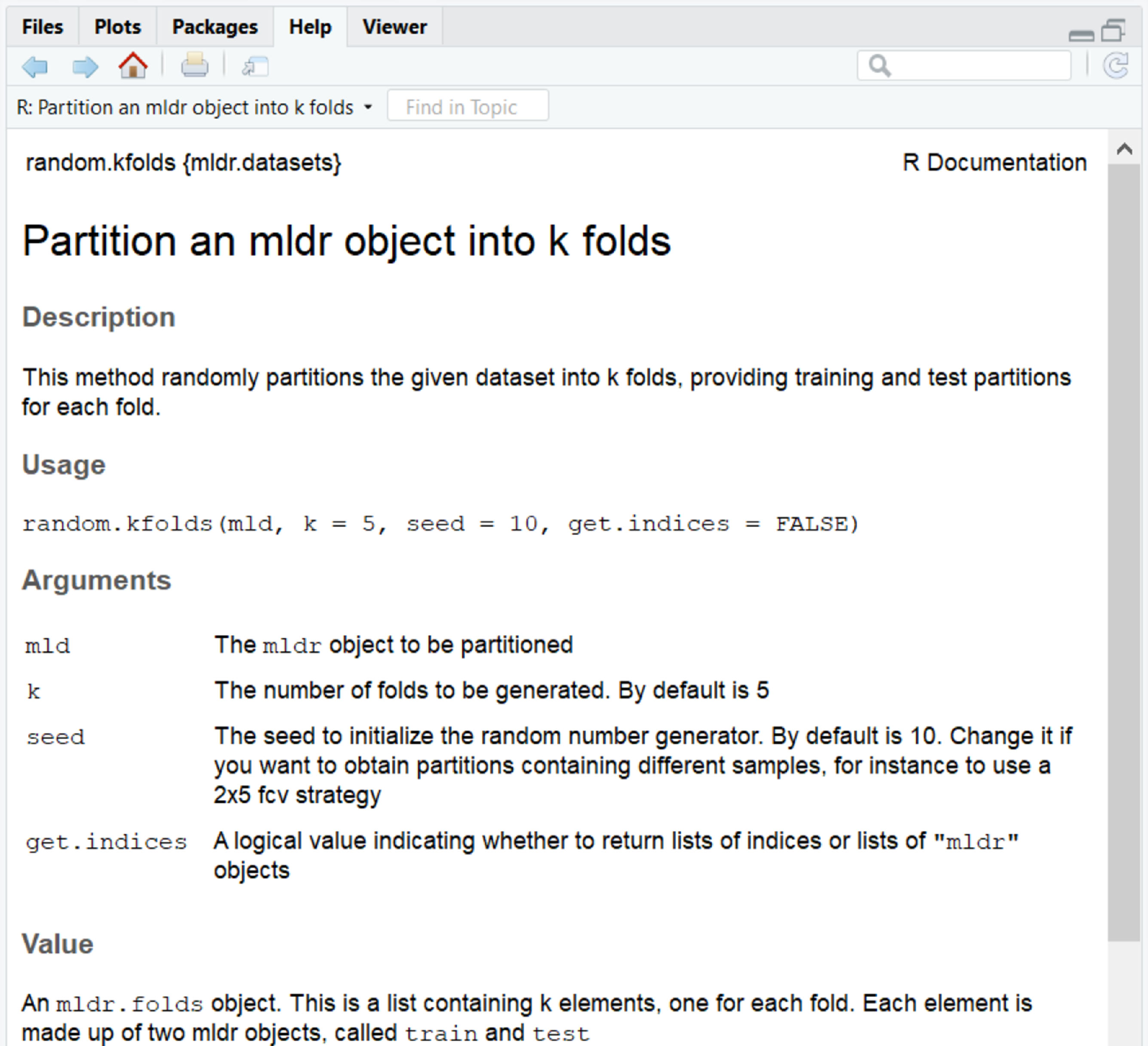}}
  \caption{Help about the random.kfolds() function.}
  \label{Fig.Help2}
\end{figure}

If you need help with a particular function, you can click on its name in the previous index. You can also use the command \texttt{help("function.name")} or, if you have already entered the name of the function into the R console or editor, you can prepend a question mark to it. In all cases, a description of the function and its parameters will be obtained, as shown in Fig. \ref{Fig.Help2}.

The subsequent sections introduce most of the available functions in the \texttt{mldr.datasets} package, showing how they can be used to perform each type of task. 

\subsection{How to load and import multi-label datasets}
The package includes not only the functions mentioned below, able to import MLDs from a web repository, but also a set of 10 already integrated MLDs. They are available immediately, as soon as \texttt{mldr.datasets} is loaded into memory. These MLDs are birds \cite{briggs2012acoustic}, cal500 \cite{CAL500}, emotions \cite{emotions}, flags \cite{gonccalves2013genetic}, genbase \cite{genbase}, langlog \cite{read2010scalable}, medical \cite{medical}, ng20 \cite{Lang95}, slashdot \cite{Read} and stackex\_chess \cite{QUINTA}. The following command can be entered into the R console to obtain a list of built-in datasets:

\begin{quote}
	\texttt{> data(package="mldr.datasets")}
\end{quote}

Each dataset is an R object, specifically an object of class \texttt{mldr}. This is the format defined in the homonymous package\footnote{The mldr package establishes the format of MLDs in R. It provides a user interface to ease the exploratory analysis and also performs data transformations, such as binarization and label powerset, among other functions. Please refer to \cite{Charte:mldr} for an extended description.} \cite{Charte:mldr} which, among other functionality, facilitates the reading of multi-label data in various formats and a user interface to carry out exploratory analysis of MLDs. These are implementation details that are not essential for the regular user of \texttt{mldr.datasets}. You can access these datasets simply by entering their name in the console, as you would with any R object. 

There are many other MLDs available online. Most of them are not embedded into the package, but can be downloaded and saved locally by means of the following two functions:

\begin{itemize}
	\item \texttt{available.mldrs()}: Retrieves the most up to date list of additional datasets from the Internet. This function does not need parameters. It returns as result an R \texttt{data.frame} containing the name and description, among other details, about available MLDs.
	
	\item \texttt{get.mldr()}: Loads any of the available datasets into memory, downloading it from the Internet if it were necessary. Once loaded,  users will be able to work with it as they would with any of the already built-in MLDs.
\end{itemize}

Below is an explanation on how to use these functions to complete each of the tasks associated with loading and downloading MLDs.

\subsubsection{Browsing the available datasets}
In addition to the 10 MLDs already integrated in the package, a much larger number is available online. The list of datasets is maintained and updated independently of the \texttt{mldr.datasets} software, so it can be extended in the future. The \texttt{available.mldrs()} function is in charge of obtaining the most recent list of online MLDs, returning it as a \texttt{data.frame} object.

A \texttt{data.frame} is a data structure made up of several rows (records) and columns (fields). In this case each row contains details of a dataset, while the columns provide the following data:
\begin{itemize}
	\item \texttt{Name}: The name of the MLD. The usual denomination found in the literature is used to refer to each MLD. This is the name to be given as input to the \texttt{get.mldr()} function explained below.
	
	\item \texttt{Description}: A brief description of the MLD's origin and/or nature.
	
	\item \texttt{Instances}: Number of data instances in the MLD.
	
	\item \texttt{Attributes}: Number of input attributes (features) in the MLD.
	
	\item \texttt{Labels}: Number of output attributes (labels) in the MLD.
	
	\item \texttt{URL}: Full URL from which the MLD can be downloaded. It is the address automatically used by the \texttt{get.mldr()} function to download a dataset when needed.
\end{itemize}

\begin{figure*}[ht!]
  \centering
  \fbox{\includegraphics[width=\textwidth]{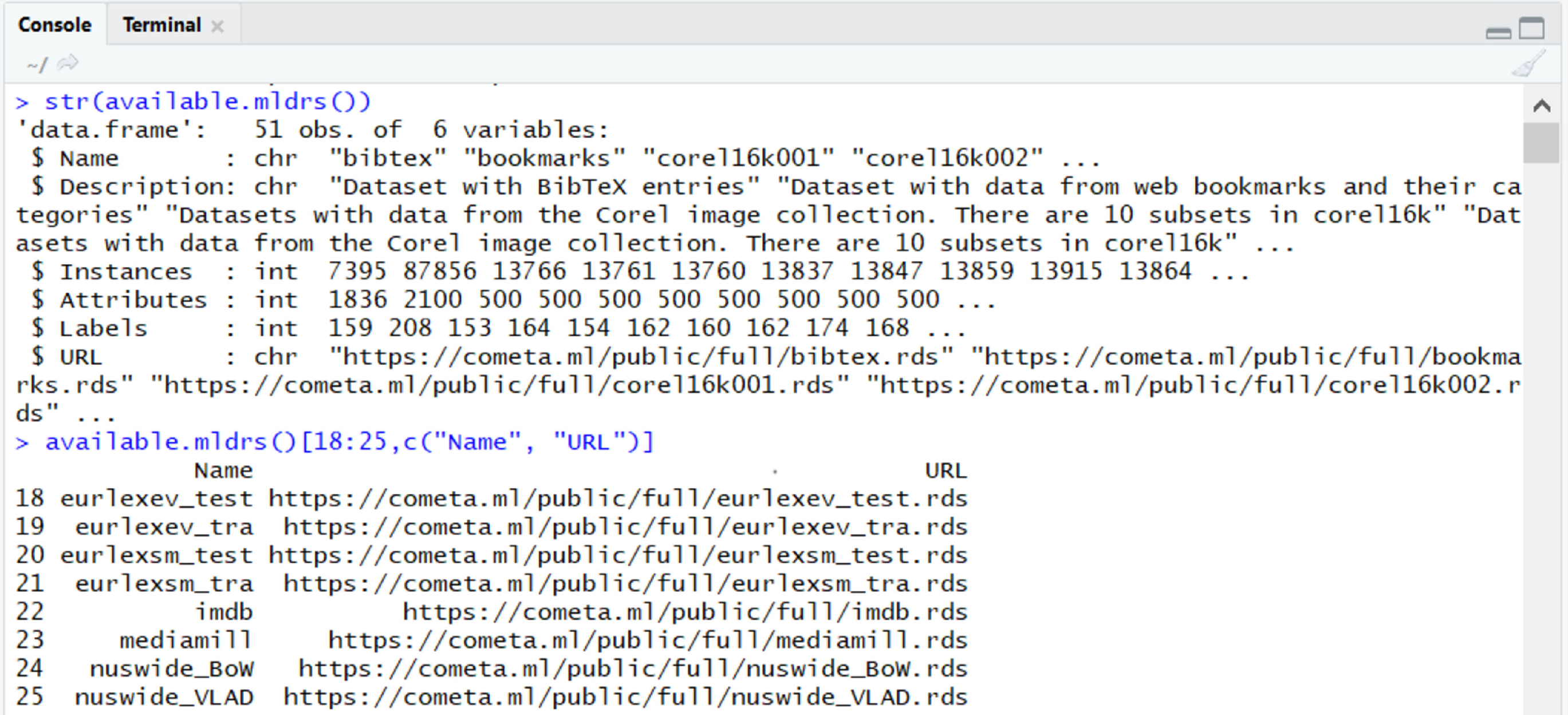}}
  \caption{Browsing the available MLDs.}
  \label{Fig.AvailableMLDs}
\end{figure*}

The result returned by the \texttt{available.mldrs()} function can be treated like any other \texttt{data.frame} in R. Fig. \ref{Fig.AvailableMLDs} shows how to get its structure, with the \texttt{str()} command (upper part), and how to recover the name and URL from some of the 60 MLDs initially available.

\subsubsection{Setting the download directory}
Before downloading any MLDs, it is possible to set the directory in which they will be stored locally. For this purpose, the environment option \texttt{mldr.download.dir} is used, whose value indicates the absolute or relative path of the desired directory. The aim is to provide a means for each user to save the downloaded MLDs where they want, so that they can load them later into memory without having to retrieve them again from the Internet. 

\begin{figure}[h!]
  \centering
  \fbox{\includegraphics[width=\columnwidth]{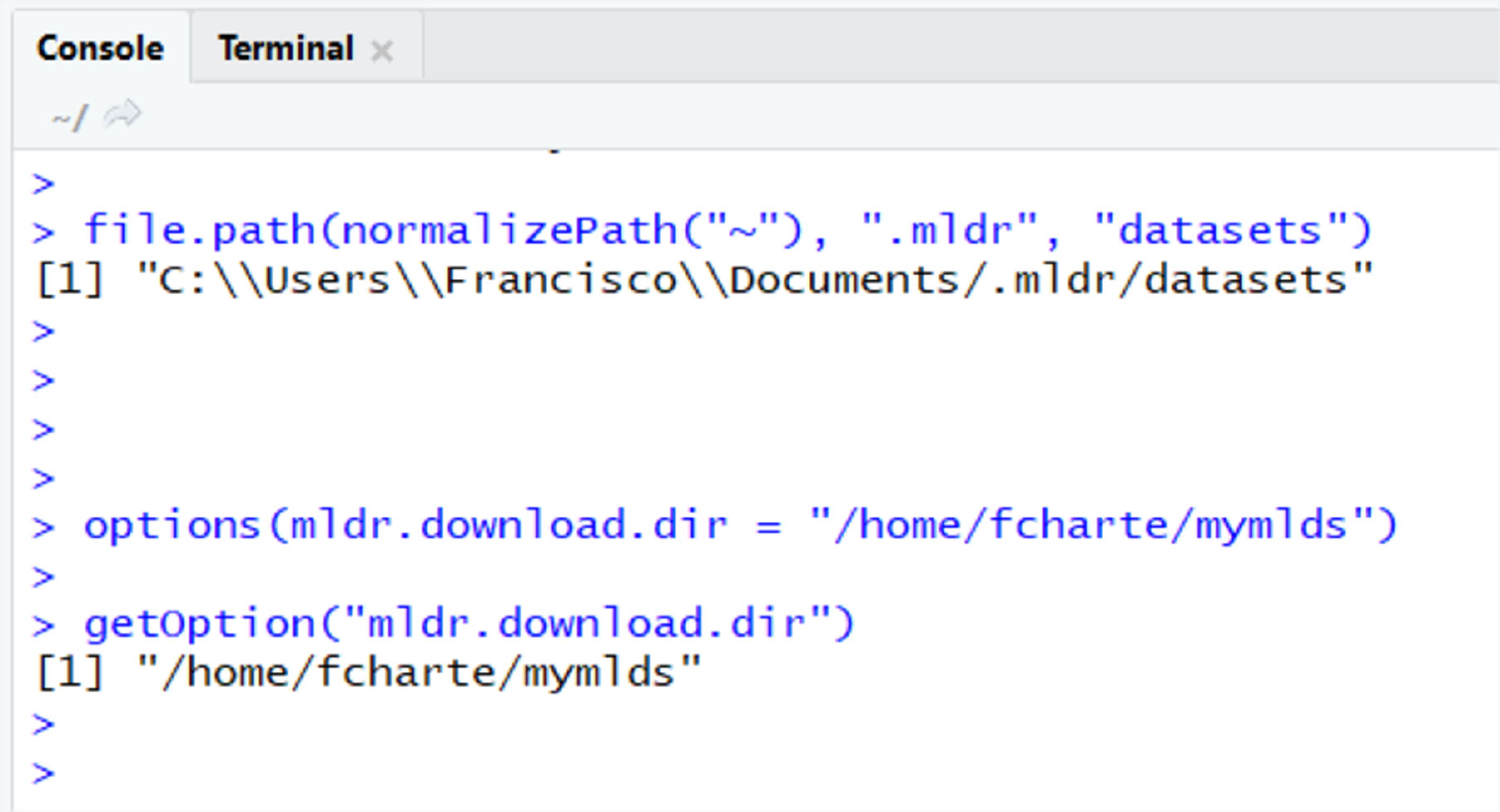}}
  \caption{Setting the directory where new MLDs will be stored.}
  \label{Fig.DownloadsDir}
\end{figure}

The \texttt{options()} command from R allows to set the value of \texttt{mldr.download.dir}. It gets one parameter, indicating the name of the variable and the value to be assigned. The bottom of Fig. \ref{Fig.DownloadsDir} shows how to do this, as well as how to check the current value through the \texttt{getOption()} command.

If a value for the previous variable is not set, and the user does not specify a download directory when invoking the \texttt{get.mldr()} function, the default download path will be used. This corresponds to a subdirectory \texttt{.mldr/datasets} that will be created in the home folder of the current user. In the case of GNU/Linux the path is usually \texttt{/home/user}, while in Windows it would correspond to the user's documents folder, as shown in the upper part of Fig. \ref{Fig.DownloadsDir}.

\subsubsection{Downloading new datasets}\label{Sec.Downloading}
In order to download an MLD you first need to know its name. It can be retrieved from the \texttt{Name} column of the \texttt{data.frame} returned by \texttt{available.mldrs()}. This name is the only parameter required to invoke the \texttt{get.mldr()} function. Optionally, the directory where you want to store the downloaded file can be specified by means of the \texttt{download.dir} parameter. If this argument is not provided by the user, the directory indicated by the option \texttt{mldr.download.dir}, as described in the previous section, will be used.

\begin{figure*}[ht!]
  \centering
  \fbox{\includegraphics[width=.9\textwidth]{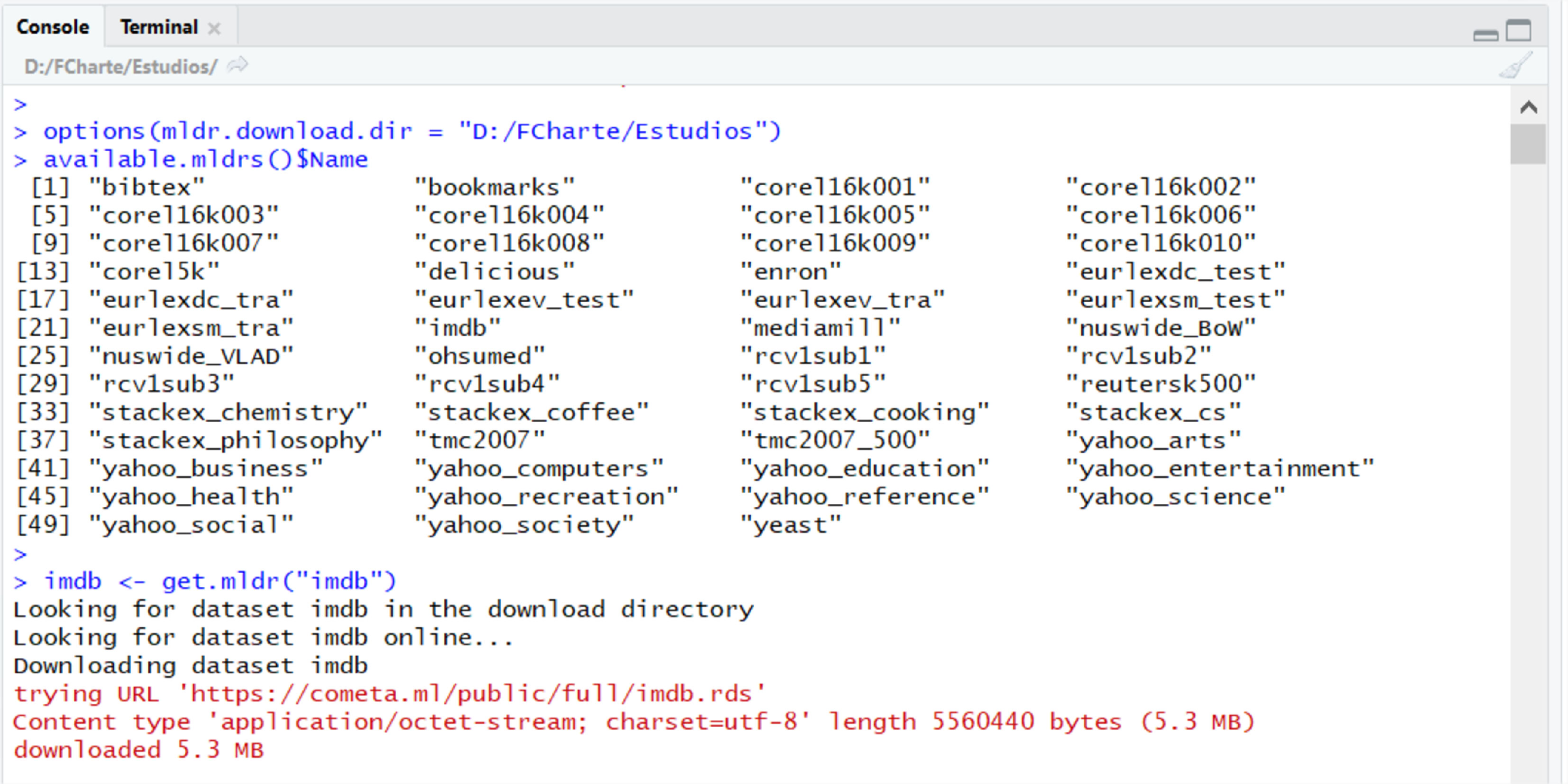}}
  \caption{Downloading a new MLD and loading it into memory.}
  \label{Fig.DownloadMLD}
\end{figure*}

The \texttt{get.mldr()} function works in three steps:
\begin{enumerate}
    \item It starts by determining the download directory. If the parameter \texttt{download.dir} has been provided by the user, this path is used, otherwise the option \texttt{mldr.download.dir} is checked. If it has been previously set, this directory is used, resorting to the default path otherwise.
    
    \item The next step is to check whether the requested MLD is already in the download directory. If this is the case, simply skip to the next step. Otherwise, it is downloaded and stored locally.
    
    \item Finally, the requested dataset is loaded into memory and returned as a result. This will be an object of class \texttt{mldr}, as explained above. It can then be used like any of the built-in MLDs.
\end{enumerate}

Fig. \ref{Fig.DownloadMLD} shows how to set the download folder, retrieve the names of all MLDs available online, and download one of them. Once the transfer is completed, the dataset is loaded into memory and stored in the \texttt{imdb} variable.

As a shortcut, in \texttt{mldr.datasets} there are individual functions defined to make it easier to download each of the MLDs available online. For example, to download the MLD \texttt{tmc2007} we can use either of the following two commands in the R console, obtaining exactly the same result.

\begin{quote}
\texttt{> tmc2007 <- get.mldr("tmc2007")}

\texttt{> tmc2007 <- tmc2007()}
\end{quote}

The operations described in the next sections are available for any \texttt{mldr} class object, be it a dataset obtained with \texttt{mldr.datasets}, generated by the \texttt{mldr} package or by any other software that produces this object format.
 
\subsection{Obtaining descriptive meta-data}\label{Sec.MetaData}
\begin{figure*}[h!]
	\centering
	\fbox{\includegraphics[width=.65\textwidth]{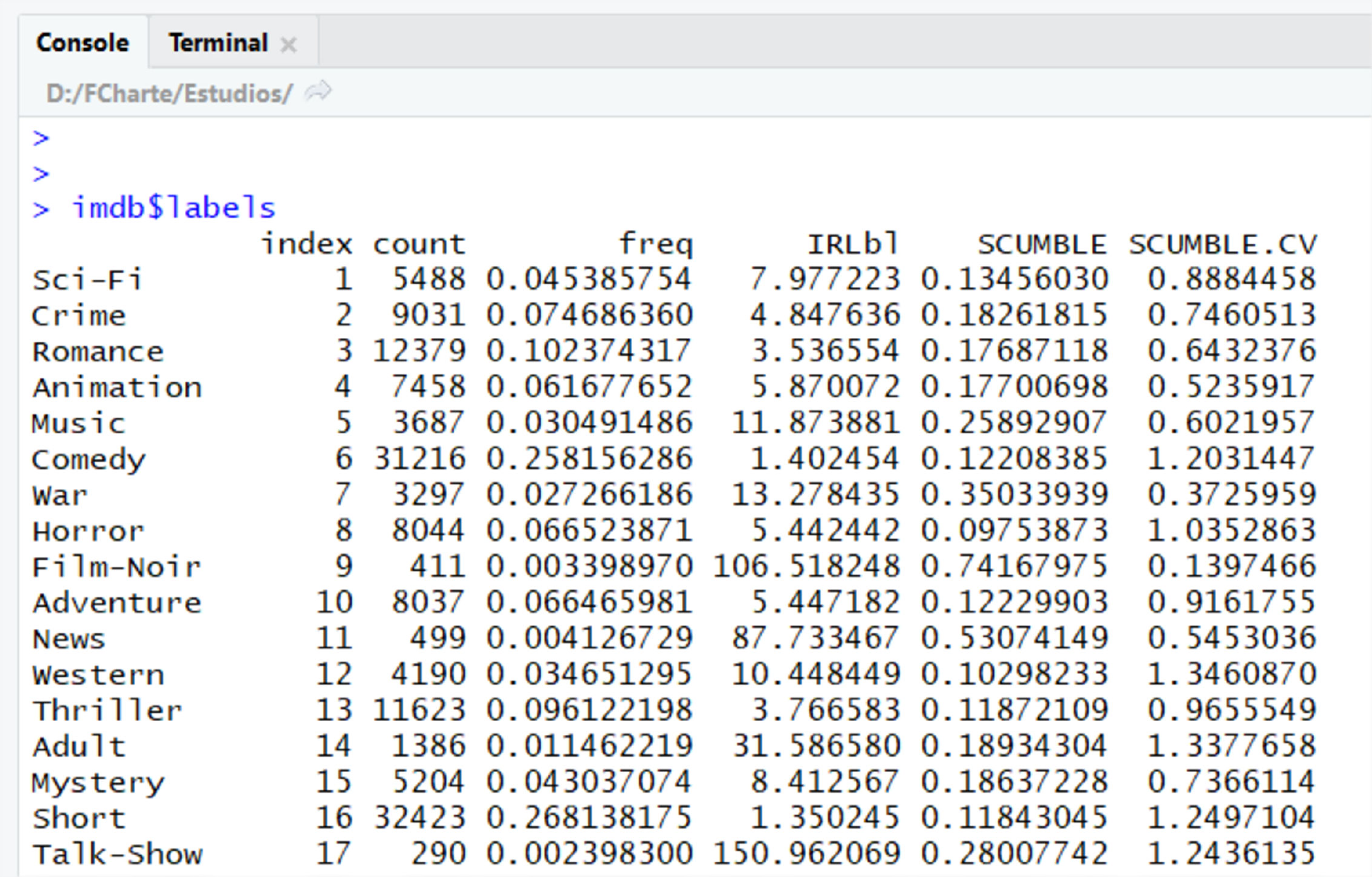}}
	\caption{List of labels in an MLD.}
	\label{Fig.Labels}
\end{figure*}

\begin{figure*}[h!]
  \centering
  \fbox{\includegraphics[width=.9\textwidth]{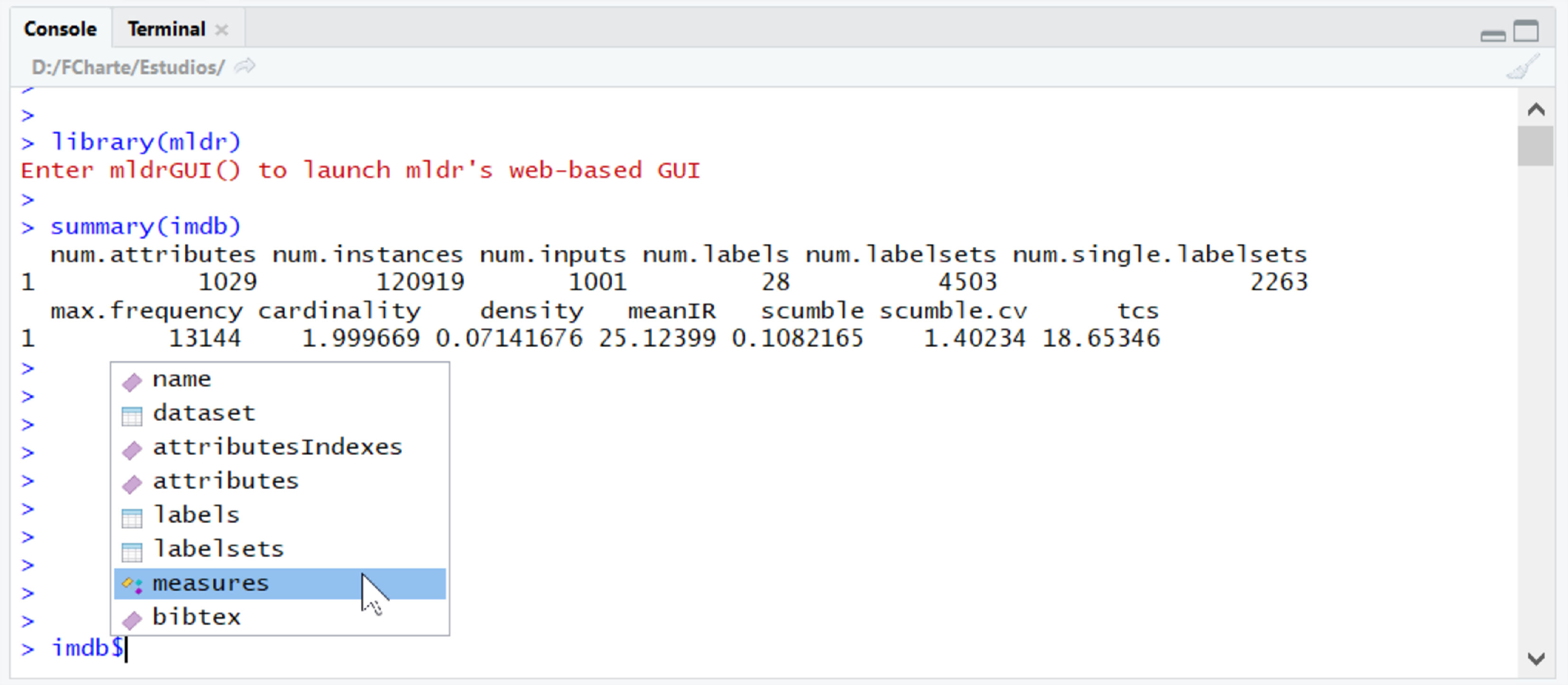}}
  \caption{Exploring the MLD traits.}
  \label{Fig.Measures}
\end{figure*}

\begin{figure*}[h!]
  \centering
  \fbox{\includegraphics[width=.9\textwidth]{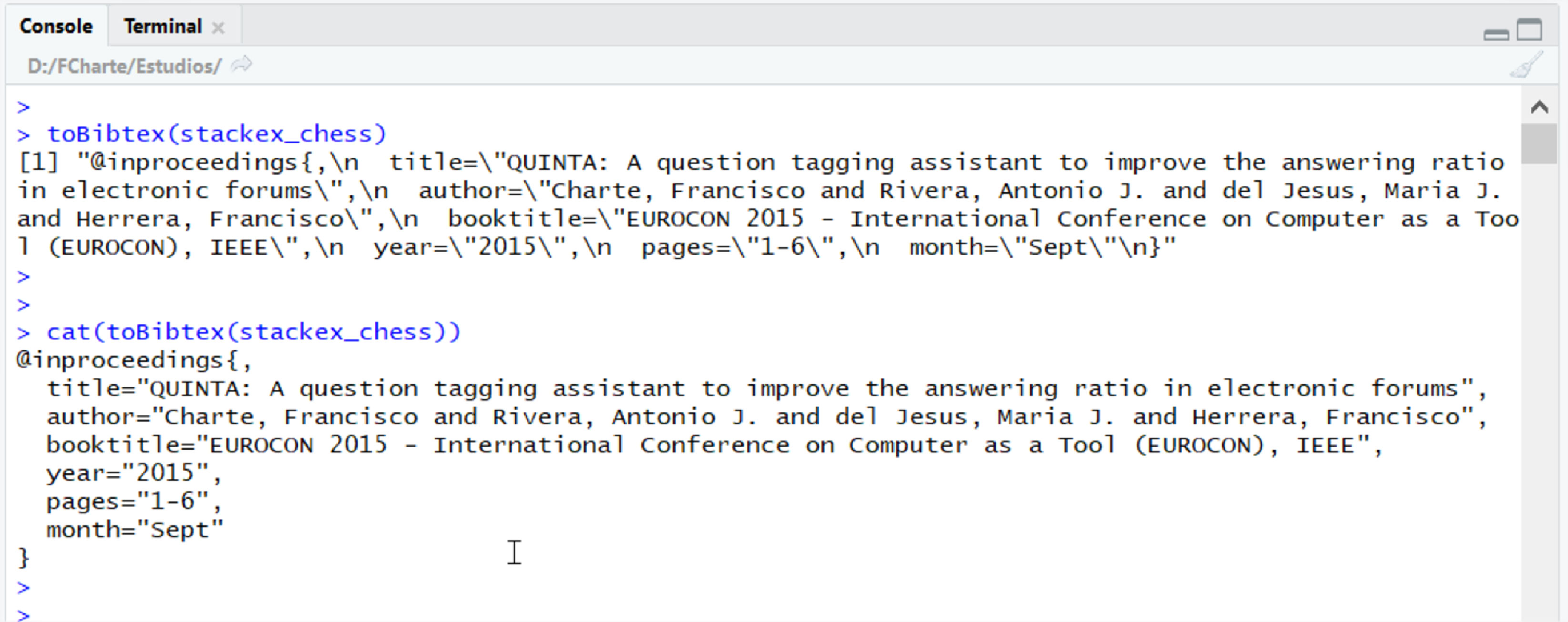}}
  \caption{Getting bibliographic data to cite a dataset.}
  \label{Fig.Bibtex}
\end{figure*}

All \texttt{mldr} objects have several fields which provide meta-data about the MLD they contain. The name and information on these fields is as follows:
\begin{itemize}
	\item \texttt{name}: Contains the original name of the MLD. Sometimes this name does not coincide with the usual name given to the dataset.
	
	\item \texttt{dataset}: A \texttt{data.frame} holding the actual data samples of the MLD.
	
	\item \texttt{attributes} and \texttt{attributesIndexes}: The former is a vector with all the MLD's attributes, including input features and output labels. For each attribute its name and data type is provided. The latter states the numeric index of input features in the MLD, since labels can be located at the beginning or at the end of the dataset.
	
	\item \texttt{labels}: It is a \texttt{data.frame} object with details about each label in the MLD (see example in Fig. \ref{Fig.Labels}), such as its name, number of occurrences, frequency, and the \textit{IRLbl} \cite{Charte:Neucom13}, \textit{SCUMBLE} and \textit{SCUMBLE.CV} \cite{Charte:NeucomSCUMBLE} metrics. 
	
	\item \texttt{labelsets}: A vector containing each label combination (\textit{labelset}) appearing in the MLD along with their number of occurrences.
	
	\item \texttt{measures}: This list provides a set of measures aimed to characterize the MLD. When the \texttt{mldr} package is loaded, the user can retrieve this list by means of the standard \texttt{summary()} command, as shown in Fig. \ref{Fig.Measures}. They can also be retrieved individually through the syntax \texttt{mld\$measures\$measureName}.
	
	\item \texttt{bibtex}: Holds the BibTeX entry needed to reference the source the MLD is coming from. This information can be also retrieved with the \texttt{toBibtex()} function. The returned string is ready to be copied to the clipboard, but can also be printed into the R console using the \texttt{cat()} command (see Fig. \ref{Fig.Bibtex}).
\end{itemize}

The \texttt{measures} list is the tool aimed to ease the selection of the proper MLDs. \highlight{Most of them were described in Section \ref{Sec.Background}}. It contains 13 different metrics:

\begin{itemize}
    \item \texttt{num.attributes}, \texttt{num.inputs} and \texttt{num.labels}: The total count of attributes in the dataset, how many of them are input features and how many output labels, respectively. The first field always is the sum of the other two.
    
    \item \texttt{num.instances}: The number of instances in the dataset.
    
    \item \texttt{num.labelsets} and \texttt{num.single.labelsets}: The former indicates how distinct labelsets appear in the MLD, while the latter states how many of them appear only once.
    
    \item \texttt{max.frequency}: Provides the number of times that the most common labelset occurs in the MLD.
    
    \item \texttt{cardinality} and \texttt{density}: These fields contain the measures known as label cardinality (\textit{Card}) and label density (\textit{Dens}).
    
    \item \texttt{meanIR}: Holds the \textit{meanIR} measure. The \textit{IRLbl} metric is stored in the homonymous column of the \texttt{labels} field described above. 
    
    \item \texttt{scumble} and \texttt{scumble.cv}: Provide the \textit{SCUMBLE} and \textit{SCUMBLE.CV} metrics for the dataset.
    
    \item \texttt{tcs}: Holds the \textit{TCS} metric, a evaluation of the theoretical complexity of the dataset.
\end{itemize}

All these metrics are automatically computed each time a new \texttt{mldr} object is created, for instance by means of the loading functions provided by the \texttt{mldr} R package. Jointly, these measures would ease for a practitioner the selection of the proper MLDs to be included in a experimental study.

\subsection{Partitioning the datasets}\label{Sec.Partitioning}
Once the proper set of MLDs has been selected, the next step usually consists in partitioning them so that some samples are used to train a model, while the remaining ones allow to test its performance. The \texttt{mldr.datasets} package provides us with the functions needed to accomplish this task. Three different partitioning strategies can be applied:

\begin{figure*}[h!]
  \centering
  \fbox{\includegraphics[width=.8\textwidth]{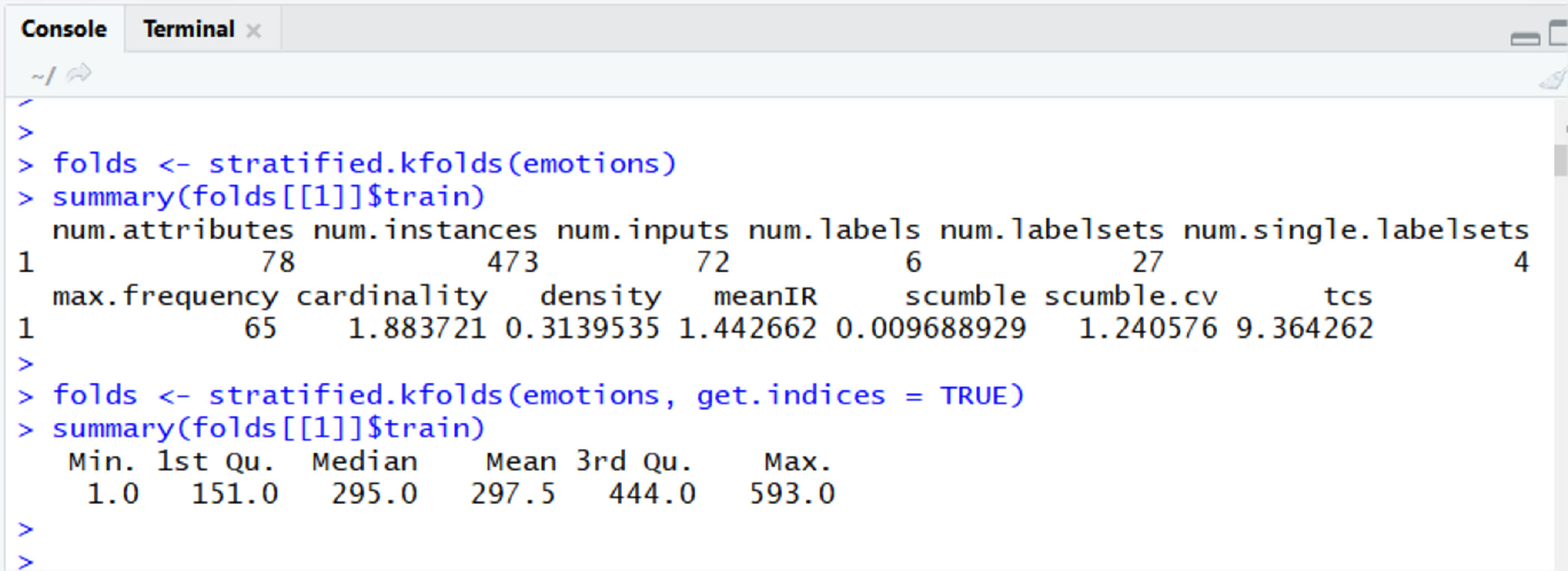}}
  \caption{Partitioning a dataset obtaining a list of mldr object (above) or a list of samples' indexes.}
  \label{Fig.Partition}
\end{figure*}

\begin{itemize}
   \item Random: Randomly separates the data samples into a certain number of partitions. As a result, the number of patterns for each label can be distributed non-uniformly. In some extreme cases, this strategy could gather all the instances for a label in the same partition. 
   
   \item Stratified: Follows the stratified algorithm described in \cite{Charte:HAIS16} in an attempt to distribute the data samples as evenly as possible among different partitions.
   
   \item Iterative stratification: The partitioning method introduced in \cite{Sechidis:2011} has the same goal of the previous one, but it tackles the problem iteratively.
\end{itemize}

For each one of these strategies, there exist three functions in the package. The first and more generic family of functions allows to create any amount of partitions with a given distribution of instances. The second group builds two partitions in a hold-out manner, one for training and one for test. The last family of functions generates partitions oriented to performing k-fold cross validation. In total, users can access the following 9 functions:
\begin{itemize}\setlength\itemsep{-.1em}
\item \texttt{random.partitions}
\item \texttt{stratified.partitions}
\item \texttt{iterative.stratification.partitions}
\item \texttt{random.holdout}
\item \texttt{stratified.holdout}
\item \texttt{iterative.stratification.holdout}
\item \texttt{random.kfolds}
\item \texttt{stratified.kfolds}
\item \texttt{iterative.stratification.kfolds}
\end{itemize}

Regardless of which of the previous methods we use, the parameters to be given to the respective function are the same in all cases:

\begin{itemize}
	\item \texttt{mld}: The only compulsory argument is the \texttt{mldr} object containing the MLD to be partitioned.
    
    \item \texttt{r} (only for \texttt{partitions} functions): A vector indicating the percentages of instances desired in each partition. For example, a value of \texttt{c(35, 25, 40)} would indicate three partitions, with 35\%, 25\% and 40\% of instances respectively.
    
    \item \texttt{p} (only for \texttt{holdout} functions): Indicates the desired percentage of instances in the training subset. It has a default value of 60.
    
    \item \texttt{k} (only for \texttt{kfolds} functions): Indicates the desired number of folds. By default it is 5, so five different folds of training/testing samples would be produced.
    
    \item \texttt{seed}: The seed used to initialize the random generator. Its default value is 10. It should be changed if we want to obtain different sets of folds, for instance to generate two sets of 5 folds (2x5fcv).
    
    \item \texttt{get.indices}: By default the partitioning functions generate a list with \highlight{as many elements as partitions} requested. \highlight{Each element will be of class \texttt{mldr} by default for the generic and hold-out function; in the case of k-folds functions,} each element will consist of two \texttt{mldr} objects, one for training and one for testing. If we assign the \texttt{TRUE} value to this parameter, lists with the indexes of the samples will be provided instead of generating \texttt{mldr} objects with the data. These indexes can be used over the original MLD to select instances, taking up much less memory space than \highlight{several} \texttt{mldr} objects.
\end{itemize}

The example shown in Fig. \ref{Fig.Partition} partitions the same dataset twice. Firstly, a list of five objects is obtained holding two members, \texttt{train} and \texttt{test}. Each one is an \texttt{mldr} object, so the same operations previously described for this class of objects are applicable. Secondly, the same MLD is partitioned enabling the \texttt{get.indices} option. In this case the \texttt{train} and \texttt{test} members are numeric vectors rather than \texttt{mldr} objects.

\subsection{How to export data to other formats}
Although the MLDs in R format can be useful to perform exploratory analysis with the \texttt{mldr} package, or conduct some experimentation using the \texttt{mlr} package \cite{mlr:multilabel}, most users would need to export them to other formats. This is the goal of the \texttt{write.mldr()} function. Currently it is able to write the content of any \texttt{mldr} object to the following formats:

\begin{itemize}
	\item MULAN: The data is written in ARFF file format following the MULAN \cite{MULAN} multi-label standard: the labels are usually located at the end of each data row, and a separate XML file containing label names is also generated.
    
    \item MEKA: As for MULAN, the MEKA \cite{MEKA} file format is also ARFF-based. However, the number and locations of labels in the data is stated in the ARFF header itself, so a separate XML file is not needed.
    
    \item KEEL: This machine learning tool \cite{triguero2017keel} also relies on the ARFF file format, as the two previous ones. The ARFF header enumerates the attributes acting as inputs and as outputs. Therefore, the labels can be located at any position on the dataset.
    
    \item LibSVM: It is the file format used by the well-known SVM library LibSVM \cite{LibSVM}. It uses sparse representation, locating the labels at the beginning of each data row.
    
    \item CSV: In case none of the previous formats fits the user's needs, they can always export to CSV format and import the data from the tool to use. This format is the standard CSV, with the attributes and labels separated by commas with these at the end. A second CSV file with the label names is also generated.
\end{itemize}

When calling the \texttt{write.mldr()} function, an \texttt{mldr} object or the value returned by one of the partitioning functions described in the previous section must be given as the first argument. In the first case a single data file will be created (and one with the labels if applicable), while in the second case as many files as \highlight{partitions contained in} the list will be generated.

The format to export the MLD \highlight{in} is indicated via the \texttt{format} parameter. This should be a string with any of the format identifiers previously enumerated, i.e. \texttt{"KEEL"}. A vector with several formats can also be given, in which case the dataset is simultaneously written in all of them. 

Many MLDs are sparse, mainly those having hundreds or thousands of input attributes. This means that only some of these attributes have a useful value in each row, the remaining ones being 0. Writing all these zeros in a text file implies a waste of space. This is the reason to use the ARFF sparse format, far more compact for sparse MLDs. The \texttt{sparse} parameter of the \texttt{write.mldr()} function takes the \texttt{FALSE} value by default. Assigning it the \texttt{TRUE} value activates this functionality. However, it should only be used with truly sparse MLDs, otherwise it will not produce any benefit. The \texttt{sparsity()} function in \texttt{mldr.datasets} can be used to check the sparsity level of any MLD. For instance:

\begin{quote}
\texttt{> sparsity(emotions)}

\texttt{[1] 0.05834739}

\texttt{>} 

\texttt{> sparsity(stackex\_chess)}

\texttt{[1] 0.9729319}
\end{quote}

As can be seen, the \texttt{emotions} dataset has less than a 6\% of sparsity, while for \texttt{stackex\_chess} the level is above 97\%. So the former should not be written as sparse, while the latter should be.

The last parameter accepted by the \texttt{write.mldr()} function is \texttt{basename}. It is useful to set the root of the filenames when several MLDs are going to be exported, usually as a result of a previous partitioning task. The original name of the MLD, stored in the \texttt{name} attribute, is used by default. If it is not a valid name, the string \texttt{"unnamed\_mldr"} will be used instead.

\begin{figure}[h!]
  \centering
  \fbox{\includegraphics[width=.9\columnwidth]{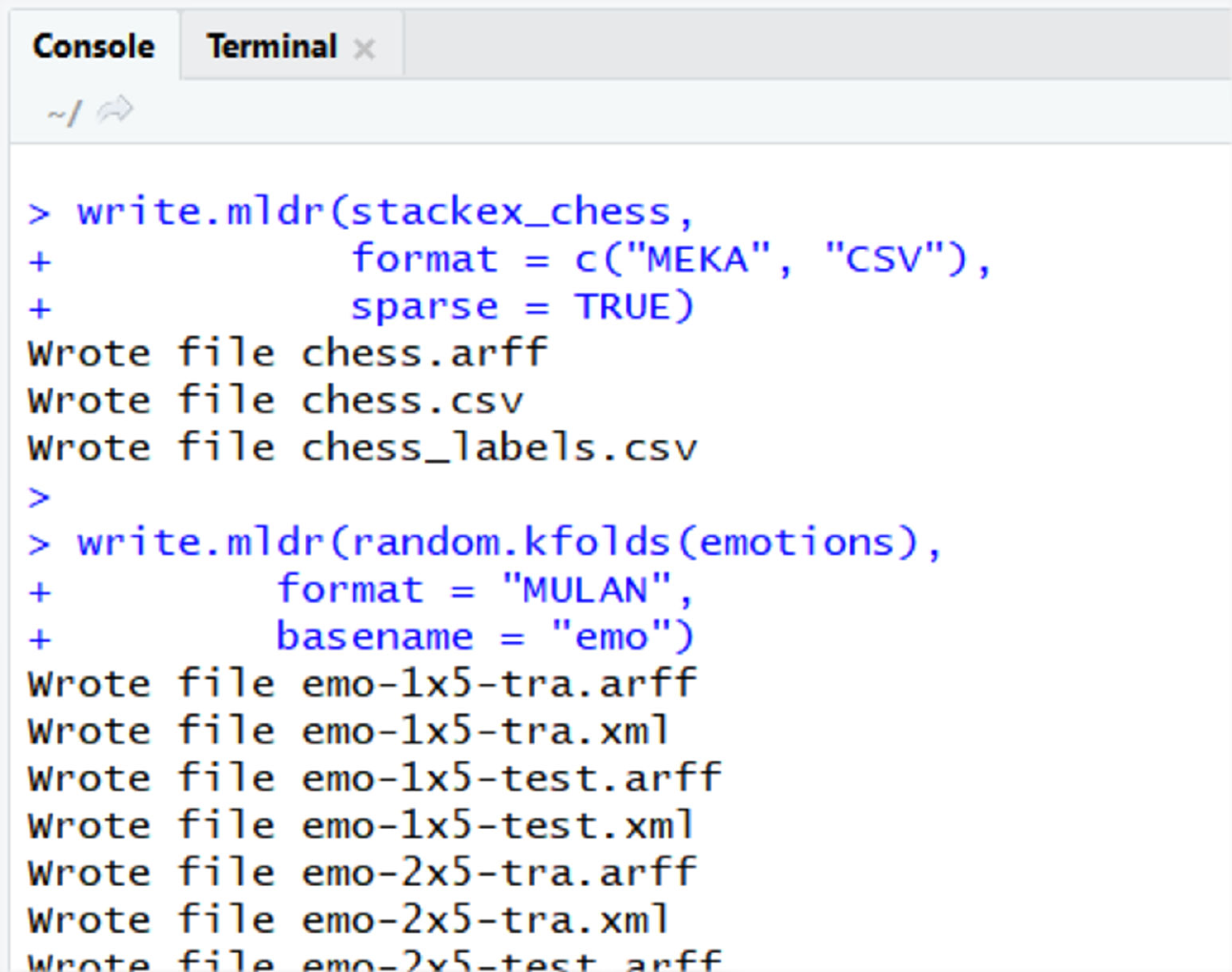}}
  \caption{Exporting a dataset and some partitions.}
  \label{Fig.Write}
\end{figure}

In Fig. \ref{Fig.Write} two typical use cases of \texttt{write.mldr()} are shown. First, a sparse MLD is written to MEKA and CSV formats. Second, a dataset is partitioned and these partitions are exported to MULAN format. 

\section{Cometa: The comprehensive multi-label data archive}\label{Sec.DataSource}
By means of the functionality offered by the package \texttt{mldr.datasets} any user can examine the characteristics of the MLDs, select the most appropriate ones for their study, partition and export them to the desired format, and reference them appropriately.  Logically, it would be desirable for such partitions to be made publicly available so that third parties can use them for comparisons. In fact, the interesting point would be that we could all use the same data partitions, thus simplifying any comparative study. This is the goal behind the comprehensive multi-label data archive (Cometa).

Using the functions described in Section \ref{Sec.mldr.datasets}, we have taken many of the publicly accessible datasets, partitioned them according to different strategies, exported them to the most popular file formats and finally designed a website that acts as a repository of all that information. The repository is accessible at \url{https://cometa.ml}. Its purpose is to make it easier for researchers to use the same data partitions when conducting multi-label studies. This section describes Cometa's structure and the steps for creating your own Cometa repository with the desired datasets.

\subsection{Browse the MLDs available at Cometa}
\begin{figure}[h!]
  \centering
  \fbox{\includegraphics[width=\columnwidth]{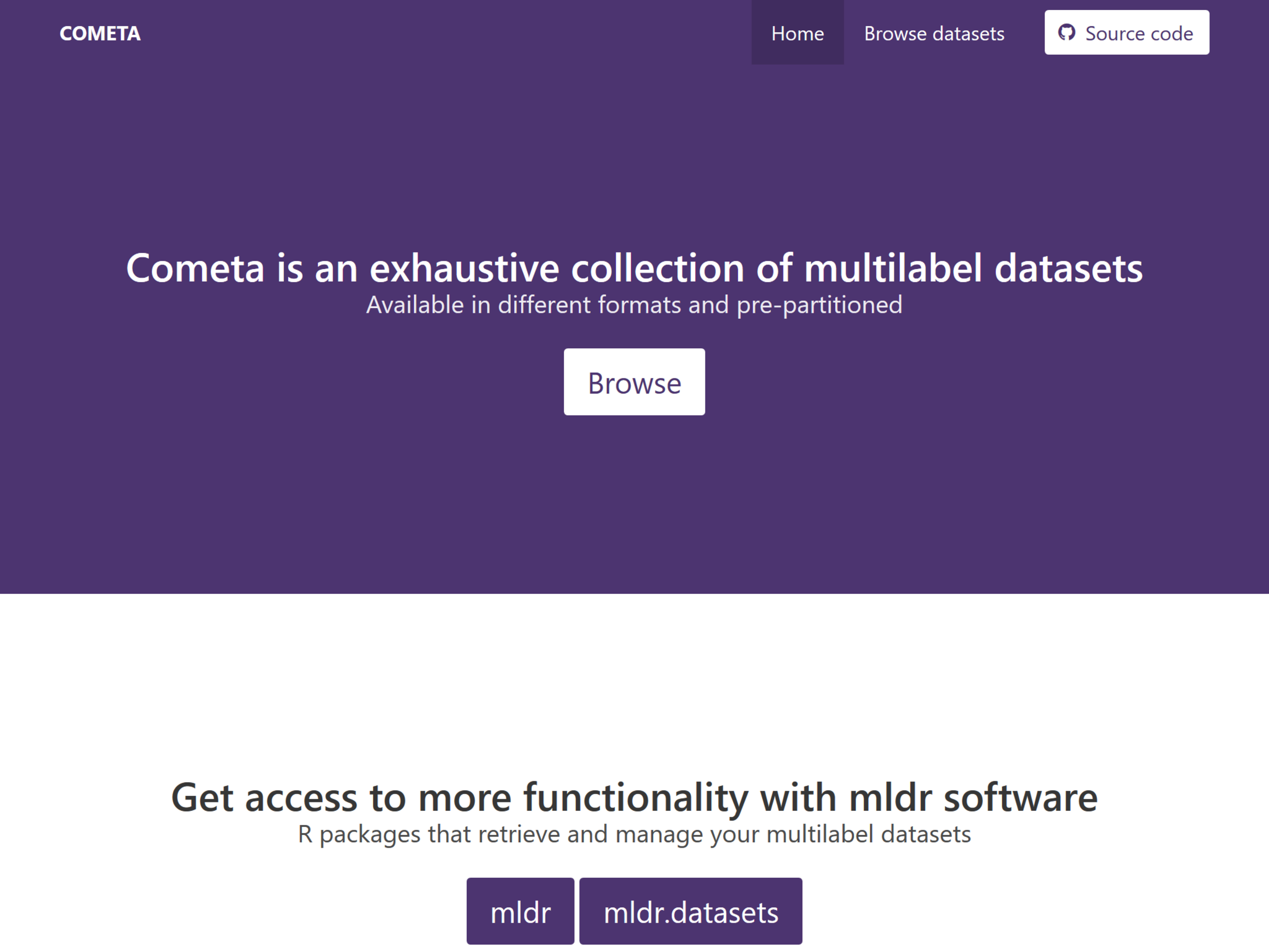}}
  \caption{Cometa main page.}
  \label{Fig.CometaMain}
\end{figure}

The main page of Cometa (Fig. \ref{Fig.CometaMain}) provides several options, aimed to ease the access to related software packages, multi-label bibliography, the source code of Cometa itself, and the list of hosted MLDs. This is accessible through the \textbf{Browse} button

\begin{table}[h!]
\setlength{\tabcolsep}{0.5em}
\def\arraystretch{0.9}
\centering
\begin{tabular}{lrcl}
\toprule
\textbf{Name} & \textbf{MLDs} & \textbf{Ref.} & \textbf{Field} \\
\midrule
 bibtex        & 1    & \cite{bibtex}              & Text \\
 birds \ding{51}         & 1   & \cite{briggs2012acoustic}  & Sound/Music \\
 bookmarks     & 1    & \cite{bibtex}              & Text \\
 cal500 \ding{51}        & 1   & \cite{CAL500}               & Sound/Music\\
 corel16k      & 10   & \cite{corel16k}    & Image \\
 corel5k       & 1   & \cite{corel5k}   & Image \\
 delicious     & 1    & \cite{HOMER}               & Text \\
 emotions \ding{51}      & 1   & \cite{emotions}            & Sound/Music \\
 enron         & 1    & \cite{enron}               & Text \\
 EUR-Lex       & 3    & \cite{mencia2008efficient} & Text \\
 flags \ding{51}         & 1   & \cite{gonccalves2013genetic}& Image \\
 foodtruck     &  1 & \cite{rivolli2017food} & Other \\
 genbase \ding{51}       & 1 & \cite{genbase}            & Protein/Genetics   \\
 imdb          & 1    & \cite{Read}                & Text \\
 langlog \ding{51}       & 1    & \cite{read2010scalable}    & Text \\
 mediamill     & 1   & \cite{mediamill}          & Video   \\
 medical \ding{51}       & 1    & \cite{medical}             & Text \\
 ng20 \ding{51}         & 1    & \cite{Lang95}              & Text \\
 nus-wide      & 2   & \cite{chua2009nus}         & Image\\
 ohsumed       & 1    & \cite{joachims1998text}    & Text \\
 rcv1v2        & 5    & \cite{lewis2004rcv1}       & Text \\
 reuters       & 1    & \cite{read2010scalable}    & Text \\
 scene         & 1   & \cite{Boutell}             & Image\\
 slashdot \ding{51}      & 1    & \cite{Read}                & Text \\
 stackexchange & 6    & \cite{QUINTA}              & Text \\
 tmc2007       & 2    & \cite{Srivastava:2005}     & Text \\
 yahoo         & 11    & \cite{ueda2002parametric} & Text \\
 yeast         & 1 & \cite{Elisseeff1}          & Protein/Genetics  \\

\bottomrule
\end{tabular}
\caption{Datasets initially available in Cometa}
\label{Table:Datasets}
\end{table}

\begin{figure*}[h!]
  \centering
  \fbox{\includegraphics[width=.85\textwidth]{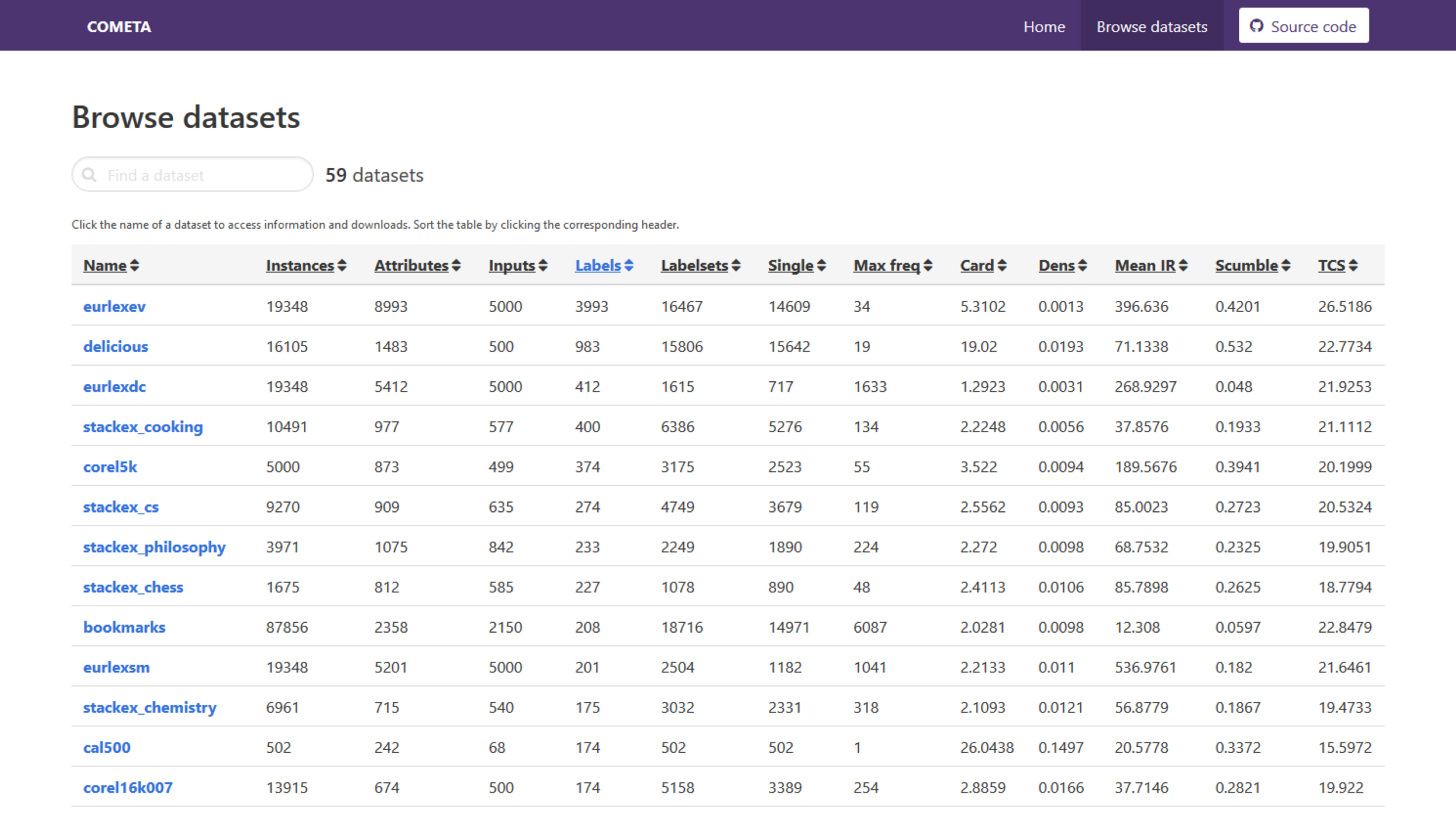}}
  \caption{Browsing the list of datasets.}
  \label{Fig.CometaDatasets}
\end{figure*}

The list of MLDs initially available in Cometa, partially visible in Fig. \ref{Fig.CometaDatasets}, is provided in Table \ref{Table:Datasets}. Those marked with a \ding{51} symbol are built-in MLDs, available as soon as the \texttt{mldr.datasets} package is loaded into memory. The remaining ones can be obtained through the \texttt{get.mldr()} function explained in subsection \ref{Sec.Downloading}.

\subsection{Filtering and searching MLDs}
The dataset browsing page provides for each MLD a set of metrics, such as the number of features, labels, labelsets, the imbalance ratio, SCUMBLE and TCS measures, etc. That list is dynamic, so that the user can change the order simply by clicking the desired column header. A second click will reverse the order. This way looking for MLDs having certain traits, i.e. those with more labels or more imbalanced, becomes a simpler process.

\subsection{Details about an MLD}
\begin{figure*}[h!]
  \centering
  \fbox{\includegraphics[trim={0 0 0 1.5cm},clip,width=.85\textwidth]{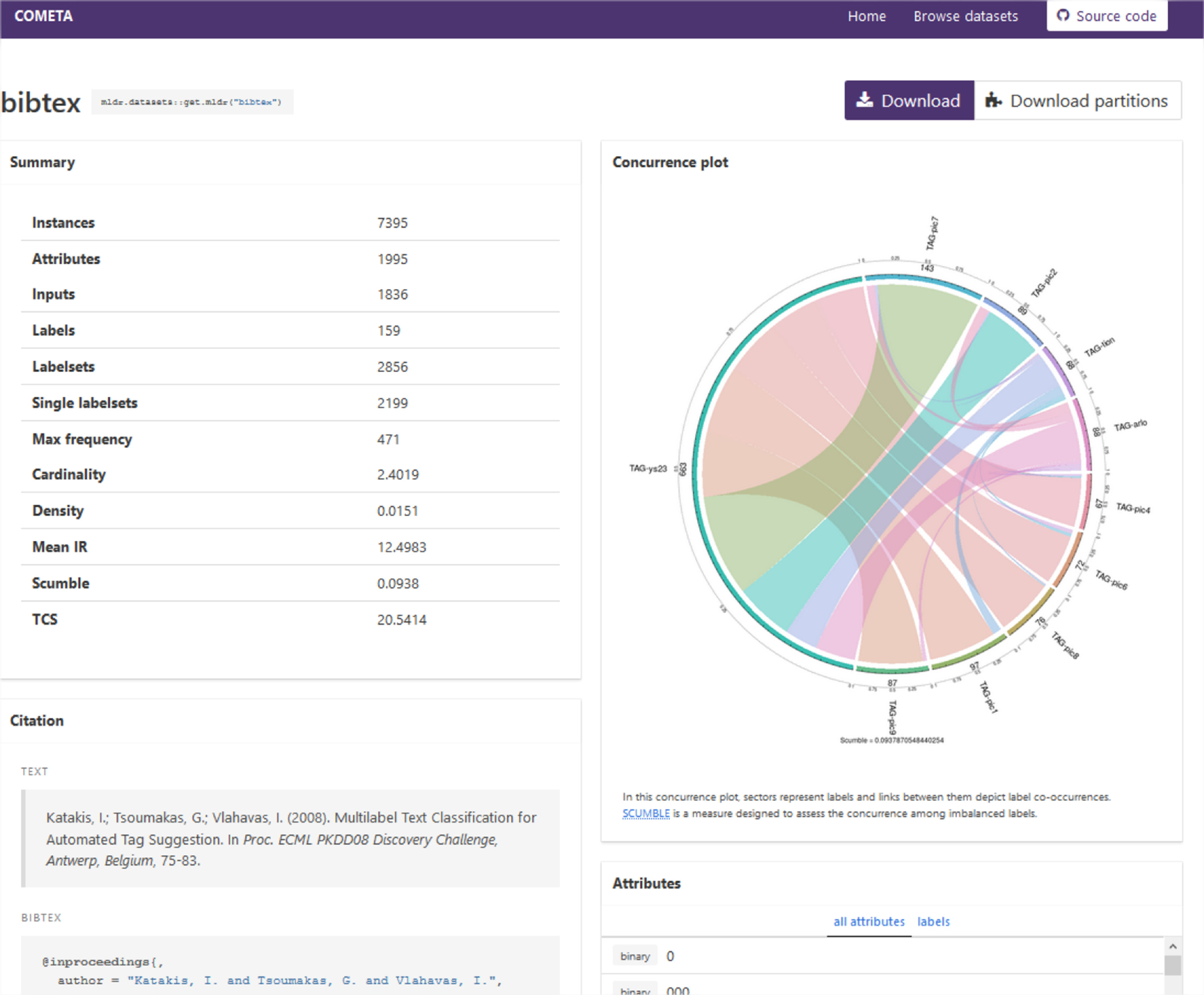}}
  \caption{MLD details page.}
  \label{Fig.CometaDetails}
\end{figure*}

In order to search a specific MLD or set of MLDs, all the user has to do is enter part of its name in the text box located at the top-left of the page. It is also valid to introduce a known value for any of the metrics shown in the list. The rows in it will be filtered as a result, easing the selection of the searched dataset.

A click on the name of any of the datasets displays its detail page (see Fig. \ref{Fig.CometaDetails}). This is composed of several panels, showing the measures that characterize the MLD, information on label concurrency, including a plot\footnote{This kind of plot, among others, can be easily generated by the \texttt{mldr} package.}, and imbalance levels, the complete list of attributes indicating their type, the list of labels and finally, the source information necessary to reference it. This additional information should be useful to decide if the MLD is appropriate for the study at glance or not. 

At the bottom of the page, a link provides all the information displayed on this page in JSON \cite{JSON} format. This facilitates the automated treatment of meta-data related to MLDs.

\subsection{Downloading data partitions}
All the datasets available at Cometa can be downloaded from R through the \texttt{mldr.datasets} package. The tip located at the top-left of the details page, following the MLD's name, provides the command to use. This will provide the full dataset, without partitioning. This same version of the MLD can be also directly downloaded from the page, by clicking the \textbf{Download} button on the right.

\begin{figure*}[htp!]
  \centering
  \fbox{\includegraphics[width=.9\textwidth]{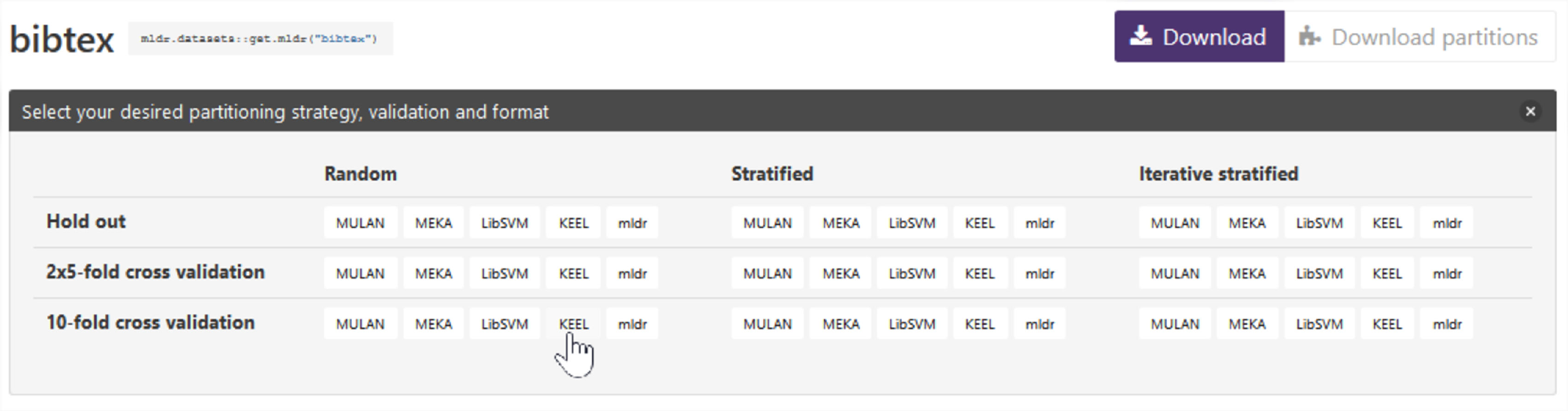}}
  \caption{Selecting the partition strategy, folds and format for downloading the MLD.}
  \label{Fig.CometaDownload}
\end{figure*}

To the right of the previous button there is another one whose objective is to allow the download of the chosen MLD already partitioned. Each of the datasets in the repository has been partitioned according to the following parameters:

\begin{itemize}
	\item \textbf{Partitioning strategy}: \highlight{All} three partitioning methods supported by \texttt{mldr.datasets} (see subsection \ref{Sec.Partitioning}) have been used with every MLD, so the user can choose between random, stratified and \highlight{iteratively} stratified partitions.
    
    \item \textbf{Number of folds}: Three configurations are provided for each partitioning strategy: hold out, 2x5-fcv and 10-fcv. The first consists of two partitions, a training partition with 60\% of the instances and a test partition with the remaining 40\%. The second one is made of two different sets of five folds, with 80\% of samples for training and 20\% for testing. The last configuration is a set of ten folds, each having 90\% of instances for training and the remaining 10\% for testing.
    
    \item \textbf{File format}: For each one of the previous nine strategy/folds configurations data have been exported to five file formats: MULAN, MEKA, KEEL, LibSVM and the mldr format. 
\end{itemize}

This adds up to a total of 45 settings for each MLD, prepared to download and use in any experimental study. All of them are accessible through the \textbf{Download partitions} button previously mentioned. It opens a window as the one shown in Fig. \ref{Fig.CometaDownload}, from where the a .tar.gz file for each case is available.

\subsection{How to host your own Cometa repository}

Since Cometa is based exclusively on open-source software, other researchers can build their own multi-label data repository by running the same software. However, installing the required software, manually partitioning the datasets and extracting their metadata might be a tedious task. In order to relieve them from this work, we provide a mostly automatic, menu-based assistant in the form of a Docker \cite{Docker} image\footnote{The Docker image is hosted on the Docker Hub at \url{https://hub.docker.com/r/fdavidcl/cometa/}}.

The assistant will automatically process datasets, but it will need them in mldr format for this. Provided that we are working in R, the \texttt{mldr()} or \texttt{mldr\_from\_dataframe()} functions from the mldr package can convert a dataset to mldr format. Afterwards, we just need to save this object into a file with the built-in function \texttt{saveRDS()}.

% ¿Pongo un ejemplo de lo que se describe en el párrafo de arriba?

Assuming now that the directory containing the public data for the repository will be located in \texttt{\textasciitilde{}/public}, we should save our datasets in RDS file format inside \texttt{\textasciitilde{}/public/full}. At this point, the Cometa assistant can run by starting the Docker image in interactive mode and forwarding one port from the host to the 80 port in the container:

\begin{small}
\begin{verbatim}
$ docker run -itp 8080:80 --mount \
  type=bind,source="~/public",target=/usr/app/public \
  fdavidcl/cometa
\end{verbatim}
\end{small}

After downloading the required image, the main menu of the Cometa assistant will show some options:

\begin{enumerate}
\item Partition datasets
\item Create summaries of your data
\item Modify website configuration
\item Launch Cometa server
\item Quit
\end{enumerate}

These options are intended to be processed in order. When choosing the first option, the assistant will scan the \texttt{public/full} folder for datasets, partition them according to different partitioning and validation strategies, and export them in a variety of formats into \texttt{public/partitions}. This process can take several hours depending on the size of the datasets, but it is only needed once. If serving dataset partitions is not desired, this option can be skipped safely.

The second option will again read the original datasets and output metadata in \texttt{public/json}. Running this option is required in order for the datasets to appear on the website. The third option will allow the user to modify parameters such as the website title or its accent color, and the fourth one will start a web server hosting the current datasets, which will be accessible at \texttt{localhost:8080}.

When partitions and metadata have been created, you may want to start the web server without requiring human interaction. This can be achieved by running Docker in detached mode:

\begin{small}
\begin{verbatim}
$ docker run -dp 8080:80 --mount \
  type=bind,source="~/public",target=/usr/app/public \
  fdavidcl/cometa
\end{verbatim}
\end{small}

After this command is run, the program will automatically build the website and serve it. That way, the server can be launched at system startup if desired.

\section{Concluding remarks}\label{Conclusions}
The use of multi-label classification algorithms is becoming increasingly widespread, given the breadth of its applications. It is therefore important to design increasingly efficient methods that are tailored to specific needs. The behavior and performance of these new methods must always be validated experimentally. This requires appropriate procedures and tools.

\highlight{
This paper attempts to help improving the way multi-label experimentation is conducted through several contributions. First, we have identified the main traps that the practitioner can find while performing multi-label experiments, and then a set of good practices has been provided. In addition, we have developed and introduced the tools needed to follow these recommendations, thus easing this kind of work.
}

In the first sections of this article we have tried to compile a set of good practices regarding how a multi-label experimentation should be conducted. According to our experience, most mistakes are due to incorrect selection or processing of MLDs. Most of the \highlight{pieces of advice provided} relate to this aspect.

Aiming to ease the usual steps followed in a multi-label experimentation, we have developed a specific software: the \texttt{mldr.datasets} R package. As has been explained in Section \ref{Sec.mldr.datasets}, the functionality provided by this software makes easier the selection, partitioning, documentation and exporting of MLDs. \texttt{mldr.datasets} is free software available to any R user, and it is open to future extensions by the authors and the community.

Even those who are not R users can benefit from the functionality of this software package, thanks to Cometa, the repository from which 60 MLDs with different partitioning strategies, number of partitions and formats can be downloaded. The main objective of this repository is to facilitate that new multi-label studies always use the same MLD partitions. This would allow future comparisons between algorithms, without the need for each researcher to re-run all results for published methods. All that would have to be done is to take the same partitions of data used in the reference article. Like \texttt{mldr.datasets}, Cometa is free software and any user can set up their own repository, as well as contribute to Cometa by providing additional datasets.

\highlight{As future work, we aim to enlarge the collection of MLDs hosted in Cometa, as well as extend the information provided for each one of them. The functionality of the \texttt{mldr.datasets} package could be also enhanced, for instance allowing any user to upload and process their datasets automatically from the R command line.}

\textbf{Acknowledgments}: \highlight{The authors are grateful to the editor and anonymous reviewers for their suggestions and advice, who have contributed to improving the content of this paper.}

This work is supported by the Spanish National Research Projects TIN2014-57251-P and TIN2015-68454-R and the Project BigDaP-TOOLS - Ayudas Fundaci\'on BBVA a Equipos de Investigaci\'on Cient\'ifica 2016.

\section*{References}

\bibliographystyle{elsarticle-num}
%\bibliography{references}

\begin{thebibliography}{10}
\expandafter\ifx\csname url\endcsname\relax
  \def\url#1{\texttt{#1}}\fi
\expandafter\ifx\csname urlprefix\endcsname\relax\def\urlprefix{URL }\fi
\expandafter\ifx\csname href\endcsname\relax
  \def\href#1#2{#2} \def\path#1{#1}\fi

\bibitem{Charte:SB-MLC}
F.~Herrera, F.~Charte, A.~J. Rivera, M.~J. del Jesus, {Multilabel
  Classification. Problem analysis, metrics and techniques}, Springer, 2016.
\newblock \href {http://dx.doi.org/10.1007/978-3-319-41111-8}
  {\path{doi:10.1007/978-3-319-41111-8}}.

\bibitem{TutorialVentura}
E.~Gibaja, S.~Ventura, A tutorial on multilabel learning, ACM Computing Surveys
  47~(3) (2015) 52:1--52:38.
\newblock \href {http://dx.doi.org/10.1145/2716262}
  {\path{doi:10.1145/2716262}}.

\bibitem{Clare}
A.~Clare, R.~D. King, Knowledge discovery in multi-label phenotype data, in:
  Proc. 5th European Conf. Principles on Data Mining and Knowl. Discovery,
  Freiburg, Germany, PKDD'01, Vol. 2168, 2001, pp. 42--53.
\newblock \href {http://dx.doi.org/10.1007/3-540-44794-6\_4}
  {\path{doi:10.1007/3-540-44794-6\_4}}.

\bibitem{Zhang2}
M.-L. Zhang, {Multilabel Neural Networks with Applications to Functional
  Genomics and Text Categorization}, IEEE Trans. Knowl. Data Eng. 18~(10)
  (2006) 1338--1351.
\newblock \href {http://dx.doi.org/10.1109/TKDE.2006.162}
  {\path{doi:10.1109/TKDE.2006.162}}.

\bibitem{Elisseeff1}
A.~Elisseeff, J.~Weston, {A Kernel Method for Multi-Labelled Classification},
  in: Advances in Neural Information Processing Systems 14, Vol.~14, MIT Press,
  2001, pp. 681--687.

\bibitem{Godbole}
S.~Godbole, S.~Sarawagi, {Discriminative Methods for Multi-Labeled
  Classification}, in: Advances in Knowl. Discovery and Data Mining, Vol. 3056,
  2004, pp. 22--30.
\newblock \href {http://dx.doi.org/10.1007/978-3-540-24775-3\_5}
  {\path{doi:10.1007/978-3-540-24775-3\_5}}.

\bibitem{Boutell}
M.~Boutell, J.~Luo, X.~Shen, C.~Brown, {Learning multi-label scene
  classification}, Pattern Recognit. 37~(9) (2004) 1757--1771.
\newblock \href {http://dx.doi.org/10.1016/j.patcog.2004.03.009}
  {\path{doi:10.1016/j.patcog.2004.03.009}}.

\bibitem{Charte:NeucomREMEDIAL}
F.~Charte, A.~Rivera, M.~del Jesus, F.~Herrera, {REMEDIAL-HwR: Tackling
  multilabel imbalance through label decoupling and data resampling
  hybridization}, Neurocomputing In press.
\newblock \href {http://dx.doi.org/10.1016/j.neucom.2017.01.118}
  {\path{doi:10.1016/j.neucom.2017.01.118}}.

\bibitem{spolaor2016systematic}
N.~Spola{\^o}r, M.~C. Monard, G.~Tsoumakas, H.~D. Lee, A systematic review of
  multi-label feature selection and a new method based on label construction,
  Neurocomputing 180 (2016) 3--15.
\newblock \href {http://dx.doi.org/10.1016/j.neucom.2015.07.118}
  {\path{doi:10.1016/j.neucom.2015.07.118}}.

\bibitem{LI-MLC}
F.~Charte, A.~Rivera, M.~del Jesus, F.~Herrera, {LI-MLC}: A label inference
  methodology for addressing high dimensionality in the label space for
  multilabel classification, Neural Networks and Learning Systems, IEEE
  Transactions on 25~(10) (2014) 1842--1854.
\newblock \href {http://dx.doi.org/10.1109/TNNLS.2013.2296501}
  {\path{doi:10.1109/TNNLS.2013.2296501}}.

\bibitem{Charte:Neucom13}
F.~Charte, A.~J. Rivera, M.~J. del Jesus, F.~Herrera, Addressing imbalance in
  multilabel classification: Measures and random resampling algorithms,
  Neurocomputing 163~(0) (2015) 3--16.
\newblock \href {http://dx.doi.org/10.1016/j.neucom.2014.08.091}
  {\path{doi:10.1016/j.neucom.2014.08.091}}.

\bibitem{Charte:NeucomSCUMBLE}
F.~Charte, A.~Rivera, M.~del Jesus, F.~Herrera, {Dealing with difficult
  minority labels in imbalanced mutilabel data sets}, Neurocomputing In press.
\newblock \href {http://dx.doi.org/10.1016/j.neucom.2016.08.158}
  {\path{doi:10.1016/j.neucom.2016.08.158}}.

\bibitem{Charte:HAIS16}
F.~Charte, A.~J. Rivera, M.~J. del Jesus, F.~Herrera, {On the Impact of Dataset
  Complexity and Sampling Strategy in Multilabel Classifiers Performance}, in:
  Proc. 11th International Conference on Hybrid Artificial Intelligent Systems,
  HAIS'16, Vol. 9648, Springer, 2016, pp. 500--511.
\newblock \href {http://dx.doi.org/10.1007/978-3-319-32034-2\_42}
  {\path{doi:10.1007/978-3-319-32034-2\_42}}.

\bibitem{Read}
J.~Read, B.~Pfahringer, G.~Holmes, E.~Frank, {Classifier chains for multi-label
  classification}, Mach. Learn. 85 (2011) 333--359.
\newblock \href {http://dx.doi.org/10.1007/s10994-011-5256-5}
  {\path{doi:10.1007/s10994-011-5256-5}}.

\bibitem{Read:2008:2}
J.~Read, B.~Pfahringer, G.~Holmes, Multi-label classification using ensembles
  of pruned sets, in: Proc. 8th IEEE Int. Conf. on Data Mining, Pisa, Italy,
  ICDM'08, 2008, pp. 995--1000.

\bibitem{RPC}
E.~H{\"u}llermeier, J.~F{\"u}rnkranz, W.~Cheng, K.~Brinker, Label ranking by
  learning pairwise preferences, Artificial Intelligence 172~(16) (2008)
  1897--1916.
\newblock \href {http://dx.doi.org/10.1016/j.artint.2008.08.002}
  {\path{doi:10.1016/j.artint.2008.08.002}}.

\bibitem{CLR}
J.~F\"{u}rnkranz, E.~H\"{u}llermeier, E.~Loza~Menc\'{\i}a, K.~Brinker,
  Multilabel classification via calibrated label ranking, Mach. Learn. 73
  (2008) 133--153.
\newblock \href {http://dx.doi.org/10.1007/s10994-008-5064-8}
  {\path{doi:10.1007/s10994-008-5064-8}}.

\bibitem{ClusteringSurvey2015}
D.~Xu, Y.~Tian, A comprehensive survey of clustering algorithms, Annals of Data
  Science 2~(2) (2015) 165--193.
\newblock \href {http://dx.doi.org/10.1007/s40745-015-0040-1}
  {\path{doi:10.1007/s40745-015-0040-1}}.

\bibitem{Zhang4}
M.-L. Zhang, {Ml-rbf : RBF Neural Networks for Multi-label Learning}, Neural
  Process. Lett. 29 (2009) 61--74.
\newblock \href {http://dx.doi.org/10.1007/s11063-009-9095-3}
  {\path{doi:10.1007/s11063-009-9095-3}}.

\bibitem{HOMER}
G.~Tsoumakas, I.~Katakis, I.~Vlahavas, {Effective and Efficient Multilabel
  Classification in Domains with Large Number of Labels}, in: Proc. ECML/PKDD
  Workshop on Mining Multidimensional Data, Antwerp, Belgium, MMD'08, 2008, pp.
  30--44.

\bibitem{MLC-Clustering}
T.~Megano, K.~ichi Fukui, M.~Numao, S.~Ono, Evolutionary multi-objective
  distance metric learning for multi-label clustering, 2015 IEEE Congress on
  Evolutionary Computation (CEC) (2015) 2945--2952\href
  {http://dx.doi.org/10.1109/CEC.2015.7257255}
  {\path{doi:10.1109/CEC.2015.7257255}}.

\bibitem{DensityClustering}
C.~Braune, S.~Besecke, R.~Kruse, Density Based Clustering: Alternatives to
  DBSCAN, Springer International Publishing, Cham, 2015, pp. 193--213.
\newblock \href {http://dx.doi.org/10.1007/978-3-319-09259-1\_6}
  {\path{doi:10.1007/978-3-319-09259-1\_6}}.

\bibitem{DBSCAN}
M.~Ester, H.-P. Kriegel, J.~Sander, X.~Xu, A density-based algorithm for
  discovering clusters a density-based algorithm for discovering clusters in
  large spatial databases with noise, in: Proceedings of the Second
  International Conference on Knowledge Discovery and Data Mining, KDD'96, AAAI
  Press, 1996, pp. 226--231.

\bibitem{MLC-Hierarchical}
N.~Cesa-Bianchi, C.~Gentile, L.~Zaniboni, Incremental algorithms for
  hierarchical classification, Journal of Machine Learning Research 7~(Jan)
  (2006) 31--54.

\bibitem{Zhang1}
M.~Zhang, Z.~Zhou, {ML-KNN: A lazy learning approach to multi-label learning},
  Pattern Recognit. 40~(7) (2007) 2038--2048.
\newblock \href {http://dx.doi.org/10.1016/j.patcog.2006.12.019}
  {\path{doi:10.1016/j.patcog.2006.12.019}}.

\bibitem{DeepNNsArchitectures}
W.~Liu, Z.~Wang, X.~Liu, N.~Zeng, Y.~Liu, F.~E. Alsaadi, A survey of deep
  neural network architectures and their applications, Neurocomputing 234
  (2017) 11 -- 26.
\newblock \href {http://dx.doi.org/10.1016/j.neucom.2016.12.038}
  {\path{doi:10.1016/j.neucom.2016.12.038}}.

\bibitem{ReadDBN}
J.~Read, F.~P\'erez-Cruz, Deep learning for multi-label classification, arXiv
  preprint arXiv:1502.05988 abs/1502.05988.

\bibitem{DL-MLC-Land}
K.~Karalasa, G.~Tsagkatakisb, M.~Zervakisa, P.~Tsakalidesa, Deep learning for
  multi-label land cover classification, in: SPIE Remote Sensing, International
  Society for Optics and Photonics, 2015, pp. 96430Q--96430Q.
\newblock \href {http://dx.doi.org/10.1117/12.2195082}
  {\path{doi:10.1117/12.2195082}}.

\bibitem{CNN-MLC}
Y.~Wei, W.~Xia, M.~Lin, J.~Huang, B.~Ni, J.~Dong, Y.~Zhao, S.~Yan, {HCP: A
  flexible CNN framework for multi-label image classification}, IEEE
  Transactions on Pattern Analysis and Machine Intelligence 38~(9) (2016)
  1901--1907.
\newblock \href {http://dx.doi.org/10.1109/TPAMI.2015.2491929}
  {\path{doi:10.1109/TPAMI.2015.2491929}}.

\bibitem{Charte:AEReview}
D.~Charte, F.~Charte, S.~Garc{\'i}a, M.~J. del Jesus, F.~Herrera, A practical
  tutorial on autoencoders for nonlinear feature fusion: Taxonomy, models,
  software and guidelines, Information Fusion 44 (2018) 78 -- 96.
\newblock \href {http://dx.doi.org/10.1016/j.inffus.2017.12.007}
  {\path{doi:10.1016/j.inffus.2017.12.007}}.

\bibitem{Madjarov}
G.~Madjarov, D.~Kocev, D.~Gjorgjevikj, S.~D\v{z}eroski, {An extensive
  experimental comparison of methods for multi-label learning}, Pattern
  Recognit. 45~(9) (2012) 3084 -- 3104.
\newblock \href {http://dx.doi.org/10.1016/j.patcog.2012.03.004}
  {\path{doi:10.1016/j.patcog.2012.03.004}}.

\bibitem{Garcia:2008}
S.~Garc{\i}a, F.~Herrera, An extension on statistical comparisons of
  classifiers over multiple data sets for all pairwise comparisons, J. Mach.
  Learn. Res. 9~(2677-2694) (2008) 66.

\bibitem{Salva:INS2010}
S.~Garc\'ia, A.~Fern\'andez, J.~Luengo, F.~Herrera, Advanced nonparametric
  tests for multiple comparisons in the design of experiments in computational
  intelligence and data mining: Experimental analysis of power, Inf. Sciences
  180~(10) (2010) 2044 -- 2064.
\newblock \href {http://dx.doi.org/10.1016/j.ins.2009.12.010}
  {\path{doi:10.1016/j.ins.2009.12.010}}.

\bibitem{Charte:RUMDR}
F.~Charte, D.~Charte, A.~J. Rivera, M.~J. del Jesus, F.~Herrera, {R Ultimate
  Multilabel Dataset Repository}, in: Proc. 11th International Conference on
  Hybrid Artificial Intelligent Systems, HAIS'16, Vol. 9648, Springer, 2016,
  pp. 487--499.
\newblock \href {http://dx.doi.org/10.1007/978-3-319-32034-2\_41}
  {\path{doi:10.1007/978-3-319-32034-2\_41}}.

\bibitem{RProject}
{R Core Team}, \href{http://www.R-project.org/}{R: A Language and Environment
  for Statistical Computing}, R Foundation for Statistical Computing, Vienna,
  Austria (2014).
\newline\urlprefix\url{http://www.R-project.org/}

\bibitem{devtools}
H.~Wickham, W.~Chang,
  \href{http://CRAN.R-project.org/package=devtools}{{devtools}: Tools to Make
  Developing {R} Packages Easier}, r package version 1.8.0 (2015).
\newline\urlprefix\url{http://CRAN.R-project.org/package=devtools}

\bibitem{briggs2012acoustic}
F.~Briggs, B.~Lakshminarayanan, L.~Neal, X.~Z. Fern, R.~Raich, S.~J.~K. Hadley,
  A.~S. Hadley, M.~G. Betts, Acoustic classification of multiple simultaneous
  bird species: A multi-instance multi-label approach, The Journal of the
  Acoustical Society of America 131~(6) (2012) 4640--4650.
\newblock \href {http://dx.doi.org/10.1121/1.4707424}
  {\path{doi:10.1121/1.4707424}}.

\bibitem{CAL500}
D.~Turnbull, L.~Barrington, D.~Torres, G.~Lanckriet, {Semantic Annotation and
  Retrieval of Music and Sound Effects}, IEEE Audio, Speech, Language Process.
  16~(2) (2008) 467--476.
\newblock \href {http://dx.doi.org/10.1109/TASL.2007.913750}
  {\path{doi:10.1109/TASL.2007.913750}}.

\bibitem{emotions}
A.~Wieczorkowska, P.~Synak, Z.~Ra\'{s}, {Multi-Label Classification of Emotions
  in Music}, in: Intelligent Information Processing and Web Mining, Vol.~35 of
  AISC, 2006, Ch.~30, pp. 307--315.
\newblock \href {http://dx.doi.org/10.1007/3-540-33521-8\_30}
  {\path{doi:10.1007/3-540-33521-8\_30}}.

\bibitem{gonccalves2013genetic}
E.~C. Gon{\c{c}}alves, A.~Plastino, A.~A. Freitas, A genetic algorithm for
  optimizing the label ordering in multi-label classifier chains, in: Proc.
  25th IEEE International Conference on Tools with Artificial Intelligence
  (ICTAI13), 2013, pp. 469--476.
\newblock \href {http://dx.doi.org/10.1109/ICTAI.2013.76}
  {\path{doi:10.1109/ICTAI.2013.76}}.

\bibitem{genbase}
S.~Diplaris, G.~Tsoumakas, P.~Mitkas, I.~Vlahavas, {Protein Classification with
  Multiple Algorithms}, in: Proc. 10th Panhellenic Conference on Informatics,
  Volos, Greece, PCI'05, 2005, pp. 448--456.
\newblock \href {http://dx.doi.org/10.1007/11573036\_42}
  {\path{doi:10.1007/11573036\_42}}.

\bibitem{read2010scalable}
J.~Read, Scalable multi-label classification, Ph.D. thesis, University of
  Waikato (2010).

\bibitem{medical}
K.~Crammer, M.~Dredze, K.~Ganchev, P.~P. Talukdar, S.~Carroll, {Automatic Code
  Assignment to Medical Text}, in: Proc. Workshop on Biological, Translational,
  and Clinical Language Processing, Prague, Czech Republic, BioNLP'07, 2007,
  pp. 129--136.

\bibitem{Lang95}
K.~Lang, Newsweeder: Learning to filter netnews, in: Proc. 12th International
  Conference on Machine Learning, 1995, pp. 331--339.

\bibitem{QUINTA}
F.~Charte, A.~J. Rivera, M.~J. del Jesus, F.~Herrera, {QUINTA: A question
  tagging assistant to improve the answering ratio in electronic forums}, in:
  EUROCON 2015 - International Conference on Computer as a Tool (EUROCON),
  IEEE, 2015, pp. 1--6.
\newblock \href {http://dx.doi.org/10.1109/EUROCON.2015.7313677}
  {\path{doi:10.1109/EUROCON.2015.7313677}}.

\bibitem{Charte:mldr}
F.~Charte, D.~Charte, Working with multilabel datasets in {R}: The mldr
  package, The R Journal 7~(2) (2015) 149--162.

\bibitem{Sechidis:2011}
K.~Sechidis, G.~Tsoumakas, I.~Vlahavas, On the stratification of multi-label
  data, in: Machine Learnig and Knowledge Discovery in Databases, Springer,
  2011, pp. 145--158.
\newblock \href {http://dx.doi.org/10.1007/978-3-642-23808-6\_10}
  {\path{doi:10.1007/978-3-642-23808-6\_10}}.

\bibitem{mlr:multilabel}
P.~Probst, Q.~Au, G.~Casalicchio, C.~Stachl, B.~Bischl, Multilabel
  classification with r package mlr, The R Journal 9~(1) (2017) 352--369.

\bibitem{MULAN}
G.~Tsoumakas, E.~S. Xioufis, J.~Vilcek, I.~Vlahavas, {MULAN: A Java Library for
  Multi-Label Learning}, J. Mach. Learn. Res. 12 (2011) 2411--2414.

\bibitem{MEKA}
J.~Read, P.~Reutemann, \href{http://meka.sourceforge.net/#datasets}{{MEKA
  multi-label dataset repository}}.
\newline\urlprefix\url{http://meka.sourceforge.net/#datasets}

\bibitem{triguero2017keel}
I.~Triguero, S.~Gonz{\'a}lez, J.~M. Moyano, S.~Garc{\'\i}a, J.~Alcal{\'a}-Fdez,
  J.~Luengo, A.~Fern{\'a}ndez, M.~J. del Jes{\'u}s, L.~S{\'a}nchez, F.~Herrera,
  Keel 3.0: an open source software for multi-stage analysis in data mining,
  International Journal of Computational Intelligence Systems 10~(1) (2017)
  1238--1249.
\newblock \href {http://dx.doi.org/10.2991/ijcis.10.1.82}
  {\path{doi:10.2991/ijcis.10.1.82}}.

\bibitem{LibSVM}
C.-C. Chang, C.-J. Lin, Libsvm: A library for support vector machines, ACM
  Trans. Intell. Syst. Technol. 2~(3) (2011) 27:1--27:27.
\newblock \href {http://dx.doi.org/10.1145/1961189.1961199}
  {\path{doi:10.1145/1961189.1961199}}.

\bibitem{bibtex}
I.~Katakis, G.~Tsoumakas, I.~Vlahavas, {Multilabel Text Classification for
  Automated Tag Suggestion}, in: Proc. ECML PKDD'08 Discovery Challenge,
  Antwerp, Belgium, 2008, pp. 75--83.

\bibitem{corel16k}
K.~Barnard, P.~Duygulu, D.~Forsyth, N.~de~Freitas, D.~M. Blei, M.~I. Jordan,
  {Matching words and pictures}, J. Mach. Learn. Res. 3 (2003) 1107--1135.

\bibitem{corel5k}
P.~Duygulu, K.~Barnard, J.~de~Freitas, D.~Forsyth, {Object Recognition as
  Machine Translation: Learning a Lexicon for a Fixed Image Vocabulary}, in:
  Proc. 7th European Conf. on Computer Vision-Part IV, Copenhagen, Denmark,
  ECCV'02, 2002, pp. 97--112.
\newblock \href {http://dx.doi.org/10.1007/3-540-47979-1\_7}
  {\path{doi:10.1007/3-540-47979-1\_7}}.

\bibitem{enron}
B.~Klimt, Y.~Yang, {The Enron Corpus: A New Dataset for Email Classification
  Research}, in: Proc. ECML'04, Pisa, Italy, 2004, pp. 217--226.
\newblock \href {http://dx.doi.org/10.1007/978-3-540-30115-8\_22}
  {\path{doi:10.1007/978-3-540-30115-8\_22}}.

\bibitem{mencia2008efficient}
E.~L. Mencia, J.~F{\"u}rnkranz, Efficient pairwise multilabel classification
  for large-scale problems in the legal domain, in: Machine Learning and
  Knowledge Discovery in Databases, Springer, 2008, pp. 50--65.
\newblock \href {http://dx.doi.org/10.1007/978-3-540-87481-2\_4}
  {\path{doi:10.1007/978-3-540-87481-2\_4}}.

\bibitem{rivolli2017food}
A.~Rivolli, L.~C. Parker, A.~C. de~Carvalho, Food truck recommendation using
  multi-label classification, in: Portuguese Conference on Artificial
  Intelligence, Springer, 2017, pp. 585--596.
\newblock \href {http://dx.doi.org/10.1007/978-3-319-65340-2\_48}
  {\path{doi:10.1007/978-3-319-65340-2\_48}}.

\bibitem{mediamill}
C.~G.~M. Snoek, M.~Worring, J.~C. van Gemert, J.~M. Geusebroek, A.~W.~M.
  Smeulders, {The challenge problem for automated detection of 101 semantic
  concepts in multimedia}, in: Proc. 14th Annu. ACM Int. Conf. on Multimedia,
  Santa Barbara, CA, USA, MULTIMEDIA'06, 2006, pp. 421--430.
\newblock \href {http://dx.doi.org/10.1145/1180639.1180727}
  {\path{doi:10.1145/1180639.1180727}}.

\bibitem{chua2009nus}
T.-S. Chua, J.~Tang, R.~Hong, H.~Li, Z.~Luo, Y.~Zheng, {NUS-WIDE: a real-world
  web image database from National University of Singapore}, in: Proc. of the
  ACM international conference on image and video retrieval, ACM, 2009, p.~48.
\newblock \href {http://dx.doi.org/10.1145/1646396.1646452}
  {\path{doi:10.1145/1646396.1646452}}.

\bibitem{joachims1998text}
T.~Joachims, Text categorization with suport vector machines: Learning with
  many relevant features, in: Proc. 10th European Conference on Machine
  Learning, Springer-Verlag, 1998, pp. 137--142.
\newblock \href {http://dx.doi.org/10.1007/BFb0026683}
  {\path{doi:10.1007/BFb0026683}}.

\bibitem{lewis2004rcv1}
D.~D. Lewis, Y.~Yang, T.~G. Rose, F.~Li, {RCV1: A new benchmark collection for
  text categorization research}, The Journal of Machine Learning Research 5
  (2004) 361--397.

\bibitem{Srivastava:2005}
A.~N. Srivastava, B.~Zane-Ulman, Discovering recurring anomalies in text
  reports regarding complex space systems, in: Aerospace Conference, IEEE,
  2005, pp. 3853--3862.
\newblock \href {http://dx.doi.org/10.1109/AERO.2005.1559692}
  {\path{doi:10.1109/AERO.2005.1559692}}.

\bibitem{ueda2002parametric}
N.~Ueda, K.~Saito, Parametric mixture models for multi-labeled text, in:
  Advances in neural information processing systems, 2002, pp. 721--728.

\bibitem{JSON}
{ECMA International},
  \href{http://www.ecma-international.org/publications/files/ECMA-ST/ECMA-404.pdf}{The
  json data interchange format}.
\newline\urlprefix\url{http://www.ecma-international.org/publications/files/ECMA-ST/ECMA-404.pdf}

\bibitem{Docker}
{Docker Inc.}, \href{https://www.docker.com/}{Docker: Build, ship and run any
  app anywhere}.
\newline\urlprefix\url{https://www.docker.com/}

\end{thebibliography}

\end{document}